\documentclass[9.5pt, a4paper, copyright, twocolumn, gdm]{google}

\usepackage[numbers, sort&compress]{natbib}
\renewcommand{\titlefont}{\color{black}\normalfont\bfseries\fontsize{18}{20}\selectfont}
\renewcommand{\absfont}{\bfseries\linespread{1.15}\fontsize{10}{11.5}\selectfont}

\uselogo{} 

\title{Life-Bench: A Benchmark and Knowledge Graph Framework for Multimodal Personalization Beyond Concept Recognition}

\correspondingauthor{Xia Hu (xiahu@google.com)}

\reportnumber{} 

\usepackage{tabularx}
\usepackage{array}
\usepackage{algorithm}
\usepackage{algorithmic}
\usepackage{xspace}
\usepackage{booktabs}
\usepackage{graphicx}
\usepackage{multirow}
\usepackage{makecell} 
\usepackage{bbding}
\usepackage{arydshln} 
\usepackage{fontawesome5} 
\usepackage{subcaption}
\usepackage{enumitem}

\newcolumntype{C}[1]{>{\centering\arraybackslash}p{#1}}

\usepackage[capitalize,noabbrev]{cleveref}

\providecommand{\eg}{\emph{e.g.}\xspace}

\providecommand{\vs}{\emph{v.s.}\xspace}

\providecommand{\std}[1]{{\scriptsize\textcolor{darkgray}{\,$\pm$#1}}}

\newcommand{\lifegraph}{LifeGraph}
\newcommand{\ourdataset}{Life-Bench}
\newcommand{\nop}[1]{\ignorespaces}

\newcommand{\cameready}[1]{{#1}}
\newcommand{\notincamera}[1]{#1}
\newcommand{\runin}[1]{\noindent\textbf{#1}}

\renewcommand{\copyrightext}{\footerfont
  Accepted at WSDM 2027 Main Conference. This is an extended version with the full appendix.\\
  Licensed under CC BY 4.0. DOI: \href{https://doi.org/10.1145/3862322.3863537}{10.1145/3862322.3863537}. \\
  \textcopyright\, 2026 Google DeepMind. %
}

\begin{document}

\author[1]{Xia Hu}
\author[1]{Honglei Zhuang}
\author[1]{Brian Potetz}
\author[1]{Alireza Fathi}
\author[1]{Bo Hu}
\author[1]{Babak Samari}
\author[1]{Howard Zhou}

\affil[1]{\thepa{}}{}

\begin{abstract}

As large language models increasingly power personal assistants, users expect them to reason over multimodal life histories, from recognizing people to understanding events to aggregating patterns, yet existing benchmarks primarily target concept-level recognition. We introduce Life-Bench, a fully synthetic, human-verified multimodal benchmark of over 11,800 question–answer pairs across 10 tasks, organized by required evidence scope: concept identification, event understanding, and aggregated reasoning. 
Its photo-centric personal histories \cameready{show comparable embedding statistics to real user accounts.}
The interconnected structure of personal data invites graph-based solutions; we propose LifeGraph, a personal knowledge graph framework for structured retrieval with on-demand access to source visual evidence, showing particular promise on event and aggregated tasks. Systematic evaluation of four retrieval paradigms on Life-Bench shows that accuracy degrades sharply with evidence scope, falling below 0.40 on aggregated tasks, and no single paradigm dominates across categories. Performance beyond concept recognition remains modest across evaluated methods, establishing personalization over multimodal histories as an open challenge and Life-Bench as a testbed for future progress.
Benchmark and code: \url{https://github.com/google-deepmind/life-bench}.

\end{abstract}

\maketitle

\section{Introduction}
\label{sec:intro}


As personal assistants increasingly rely on large language models,
expectations for personalization grow beyond stylistic and preferential alignment: users now expect these systems to understand and reason over their personal history~\cite{zhang2025personalization}.
Much of this history is captured visually: photo albums and social media archives accumulate timestamped photographs chronicling social and personal life across years~\cite{dumais2003stuff}. 
User studies show people naturally pose complex queries over such records, from simple lookups to questions requiring multi-step reasoning across relationships, timelines, and visual evidence~\cite{li2025omniquery,mei2026atmbench,tan2023timelineqa}. 
The reasoning capabilities of modern vision-language models~(VLMs)~\cite{achiam2023gpt,google25gemini25,gemini35flash2026} now make it feasible to serve these complex queries over multimodal personal histories.

Answering such queries requires searching and reasoning across a user's multimodal history.
Questions like ``Who was with me last Christmas?'' or ``Where did my family go after the California trip?'' may demand resolving personal references, locating events across time, and reading visual details not captured in captions. 
Critically, these queries vary in the scope of personal context required, from concept-level knowledge of who is in a user's life and their relationships, to event-level understanding of what happened at a specific time and place, to history-level reasoning that aggregates evidence across events and time periods. 
Each successive scope compounds the challenge: more records to search, more modalities to align, and more evidence to compose into a coherent answer.

Evaluating these capabilities requires benchmarks spanning these context scopes, yet existing multimodal personalization benchmarks primarily target concept-level abilities, such as identifying personal entities and matching their appearance~\cite{alaluf2024myvlm, nguyen2024yollava, an2024mcllava,hao2025rap, das2025r2p}, with evaluation focused on recognition accuracy and preference alignment~\cite{kim2025mmpb}. 
Recent benchmarks have begun pushing toward longer-term and more comprehensive personal understanding~\cite{nie2026personavlm, bei2026memgallery,xue2026personalvcl, mei2026atmbench,yang2026hippocamp}. 
Nevertheless, comprehensive evaluation remains constrained by data availability:
real personal collections pose privacy and legal barriers to public release and typically cover only a handful of individuals~\cite{gurrin2016ntcir, mei2026atmbench,yang2026egolife, tran2026openlifelogqa}. 
Openly distributable synthetic alternatives include text-only personal records~\cite{tan2023timelineqa, du2024perltqa, zhang2026rhelm} and considerably smaller multimodal collections~\cite{bei2026memgallery,nguyen2026VisualMem}, some grounded in dialogue rather than visual records~\cite{bei2026memgallery}.
Advances in text~\cite{gpt54, gemini35flash2026} and image generation~\cite{gemini25image} now 
enables the synthesis of photo-centric personal histories organized around structured social networks and grounded in source visual evidence, at a publicly releasable scale. 

To this end, we introduce {\ourdataset}, a synthetic multimodal benchmark for personalization across concept, event, and history-level reasoning.
All data in {\ourdataset}~(images, social networks, and event histories) is synthetically generated,
enabling public release without exposing real users’ data.
To retain realism, generation is seeded with textual scenarios derived from YFCC100M photographs~\cite{thomee2016yfcc100m}; 
systematic comparison shows similar distributional characteristics to real user accounts~(Sec~\ref{subsec:dataset_validation}).
The benchmark comprises over 11,800 question-answer pairs across 10 tasks in three categories mirroring the context scopes above: Concept Identification, Event Understanding, and Aggregated Reasoning.
Underlying these tasks, the personal context comprises social networks and multimodal lifelogs \nop{of timestamped images and descriptions,} organized into 10 isolated virtual accounts (Vaccounts) to simulate individual users' archives; 
every question and image is verified through full-coverage human annotation.

Because personal archives grow beyond context limits~\cite{long2025m3agentg,tavakoli2026beam,zhang2025personalization}, we evaluate four retrieval families on
{\ourdataset}: sparse lexical~\cite{robertson2009bm25}, dense embedding~\cite{lewis2020rag}, concept-memory~\cite{das2025r2p,hao2025rap}, and graph-based~\cite{gutierrez2025hipporag2}.
Accuracy degrades systematically with evidence scope: methods achieve avg. $0.58$-$0.78$ on concept identification but drop to $0.30$-$0.57$ on event understanding and fall below $0.40$ on aggregated reasoning, exposing bottlenecks in both evidence retrieval and downstream reasoning.
No single paradigm dominates across tasks: each excels when required evidence matches what it indexes, yet none sustains its advantage as context scope broadens.
These failures trace to a structural mismatch: complex queries demand cross-modal evidence alignment, relational and temporal navigation, and synthesis over dispersed records; 
no evaluated method handles this combination consistently across evidence scopes.

This mismatch motivates approaches \nop{designed} to preserve the relational and multimodal structure of personal histories. 
To explore this direction, we propose {\lifegraph}, an end-to-end graph-based framework. 
{\lifegraph} extracts and encodes personal concepts, relations, and events from multimodal records as structured knowledge in graph, maintaining provenance links to source data.
At query time, {\lifegraph} combines graph traversal with on-demand access to source multimodal records through these links, supplying relational context and fine-grained visual details for reasoning.
On {\ourdataset}, LifeGraph achieves the best overall retrieval-based performance and leads on several event-level and aggregated tasks, demonstrating the promise of structured personal knowledge. 
Despite these gains, performance beyond concept recognition remains modest for all evaluated methods including {\lifegraph}, 
showing that personalization over multimodal histories remains an open challenge and that {\ourdataset} provides a testbed for measuring progress.

Our contributions are as follows. 
(1) {\ourdataset}, a human-verified multimodal benchmark for personalization over personal histories, spanning concept identification, event reasoning, and history-wide aggregation. 
(2) {\lifegraph}, an end-to-end framework that organizes personal knowledge as source-linked knowledge graphs, enabling structured retrieval with on-demand access to visual evidence. 
(3) A systematic evaluation showing that performance degrades with evidence scope, retrieval paradigms have complementary strengths, and both retrieval and reasoning remain bottlenecks.

\begin{figure*}[t]
  \centering
  \includegraphics[width=0.98\textwidth]{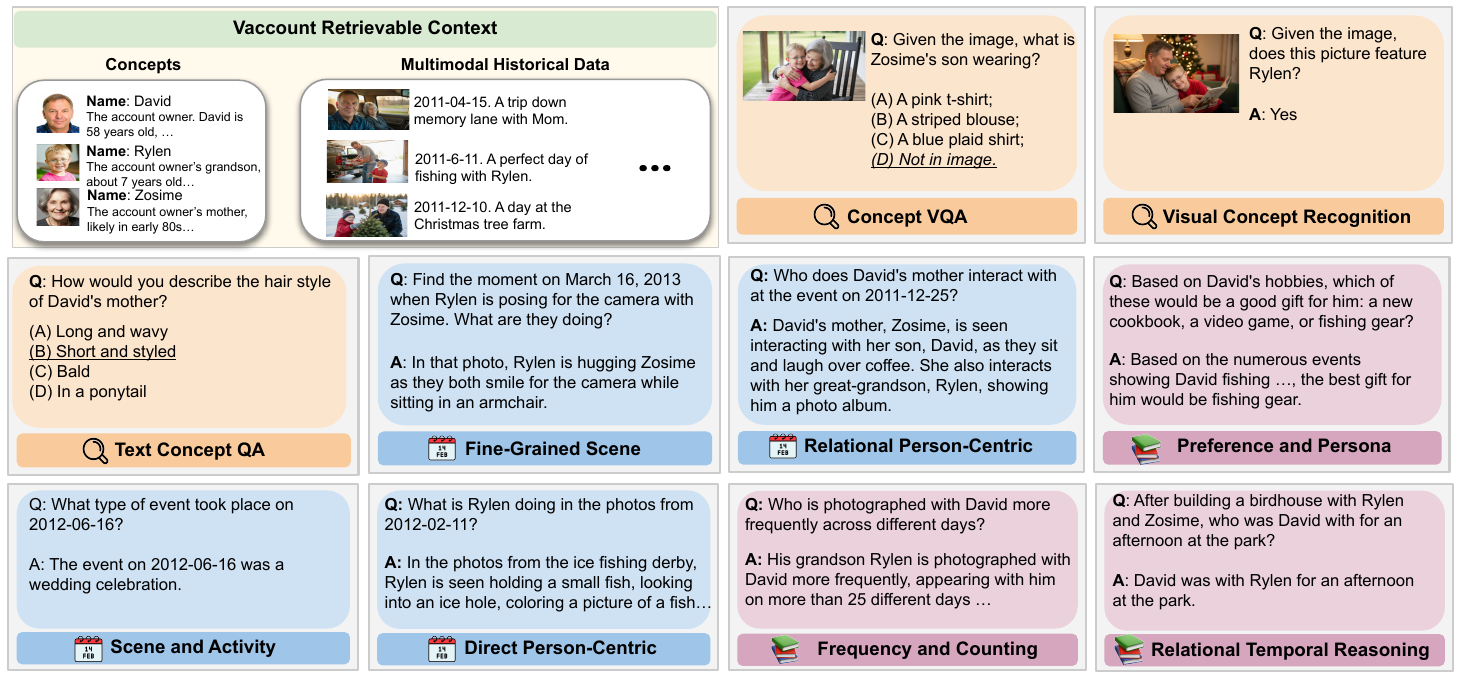}
  \caption{A {\ourdataset} Vaccount and representative examples of 10 tasks. Each Vaccount contains personal concepts and timestamped multimodal histories. Tasks are grouped by categories: Concept  Identification~(orange, \faSearch), Event Understanding~(blue, \faCalendar) and Aggregated Reasoning~(purple, \faLayerGroup).}
  \label{fig:bench_example}
\end{figure*}

\section{Related Work}
\label{sec:related}


\subsection{Personalization in LLM Era}

In large language models and autonomous agents era, expectations for personalization shifts from aligning outputs with user preferences to understanding and acting upon a user's individual context and history~\cite{zhang2025personalization}. 
Text-only studies retrieve from user-authored content for tailored generation~\cite{salemi2024lamp, kumar2024longlamp}, learn fine-grained preferences from interaction histories~\cite{zollo2025personalllm, zhao2025prefeval}, and align with diverse demographic personas~\cite{kirk2024prism, castricato2025persona}. 
In these approaches, personal data serves as a stylistic and preferential signal, matching user tendencies rather than answering queries that locate specific past events.

Complementary work builds long-term memory systems that maintain personal knowledge across sessions~\cite{packer2024memgpt, chhikara2025mem0, xu2025amem}, with existing systems and benchmarks~\cite{maharana2024locomo, wu2025longmemeval, tavakoli2026beam} focusing predominantly on conversational histories. 
Recent efforts extend beyond dialogue to emails~\cite{zhang2026rhelm} and multimodal inputs like screenshots~\cite{wang2025mirix} and video streams~\cite{long2025m3agentg}, though multimodal content is typically captioned into text before storage~\cite{liu2026memverse}, discarding fine-grained visual evidence. 
Surveys identify building memory over raw multimodal personal data as an open frontier~\cite{du2026memorysurvey, hu2026memoryagentsurvey}.

Beyond conversations, photo albums and social media archives preserve timestamped visual records of relationships, activities, and life events~\cite{dumais2003stuff}, motivating question answering over personal photo collections~\cite{jiang2025memoryqa}, lifelog retrieval~\cite{gurrin2016ntcir, tran2026openlifelogqa}, and egocentric video understanding~\cite{yang2026egolife}. 
User studies~\cite{li2025omniquery} and task designs~\cite{tan2023timelineqa, mei2026atmbench} show that natural queries over such records span from locating and reasoning over specific events to history-wide temporal reasoning, frequency counting, and cross-event composition. 
Evaluation remains constrained by data availability: real collections rarely allow public release and cover few individuals~\cite{gurrin2016ntcir, mei2026atmbench, yang2026egolife, tran2026openlifelogqa}, while open synthetic benchmarks offer either text-only records~\cite{tan2023timelineqa, du2024perltqa, zhang2026rhelm} or much smaller multimodal data~\cite{bei2026memgallery}.

Existing multimodal personalization studies largely target concept-level capabilities: recognizing personal entities via learned embeddings~\cite{alaluf2024myvlm, nguyen2024yollava, an2024mcllava} or retrieval-time matching~\cite{hao2025rap, das2025r2p},
with benchmarks emphasizing recognition and preference~\cite{kim2025mmpb} over historical reasoning.
Recent benchmarks move toward long-term personal understanding via multi-session dialogue~\cite{nie2026personavlm, bei2026memgallery} (focusing on conversations over visual records), real personal data~\cite{jiang2025memoryqa, xue2026personalvcl, mei2026atmbench}, personal document archives~\cite{yang2026hippocamp}, and synthetic photo histories~\cite{nguyen2026VisualMem}. 
Earlier, MemexQA~\cite{jiang2017memexqa} was limited to single-album lookup over 101 real users, while TimelineQA~\cite{tan2023timelineqa} introduced temporal and aggregative queries on text-only synthetic timelines. 
To our knowledge, no existing benchmark combines synthetic photo-centric personal histories with structured social networks, tasks graded by evidence scope from concept identification through event reasoning to history-wide aggregation, question-level provenance to supporting source records where applicable, and publicly releasable scale. 

\subsection{Structural and Graph-based Retrieval}

To retrieve evidence that requires navigating relationships and composing information across records, recent work organizes corpora into entity-relation graphs for structured retrieval~\cite{peng2024graphragsurvey, han2025graphrag, guo2025lightrag}.
Think-on-Graph~\cite{sun2024tog} collects relational evidence paths via beam search over a KG, and its successor jointly retrieves graph paths and linked documents~\cite{ma2025tog2};
HippoRAG~\cite{gutierrez2025hipporag2, hipporag} adapts this paradigm for long-term memory.
On the personal side, Personal Knowledge Graphs~\cite{balog2019personal, bernard2024pkg, yang2022pkg} have been built from text lifelogs~\cite{yen2019pkb} and later text-and-image inputs~\cite{yen2021personal}, 
but retrieval consistently returns text, as in LLM-era graph memory such as Graphiti~\cite{rasmussen2025zep} and general-corpus multimodal graph-RAG~\cite{wan2026mmgraphrag}.
LifeGraph extends graph-based retrieval to multimodal personal histories by organizing user-specific concepts, relations, and events in a knowledge graph linked to source records, and retrieving the original multimodal records on demand for reasoning.

\section{{\ourdataset}}
\label{sec:dataset}


\begin{table}[t]
\centering
\caption{Key statistics of {\ourdataset}.}
\label{tab:dataset_statistics}
\resizebox{0.8\linewidth}{!}{%
\begin{tabular}{lr}
\toprule
\textbf{Statistic}                        & \textbf{Number} \\ \midrule
Total Questions                           & 11,811          \\
\quad * Multiple-choice Questions         & 1,048           \\
\quad * Binary Questions                  & 1,613           \\
\quad * Open Questions                    & 9,150           \\ \midrule
Total Tasks                               & 10              \\
\quad * Concept Identification            & 3               \\
\quad * Event Understanding               & 4               \\
\quad * Aggregated Reasoning              & 3               \\ \midrule
Total Concepts                            & 33              \\
\quad * Human Concepts                    & 28              \\
\quad * Animal Concepts                   & 5               \\ \midrule
Total Events                              & 382             \\ \midrule
Total Images                              & 2,887           \\
\quad * Concept Portraits                 & 33              \\
\quad * Historical Images                 & 2,479           \\
\quad * Query Images                      & 375             \\ \midrule
Total Vaccounts                           & 10              \\
\quad * Avg. Concepts per Vaccount        & 3.3             \\
\quad * Avg. Events per Vaccount          & 38.2            \\
\quad * Avg. Images per Vaccount          & 288.7           \\
\quad * Avg. Questions per Vaccount       & 1,181.1         \\ \bottomrule
\end{tabular}%
}
\end{table}

\begin{figure*}[t]
  \centering
  \begin{subfigure}[b]{0.23\linewidth}
    \centering
    \includegraphics[width=\linewidth]{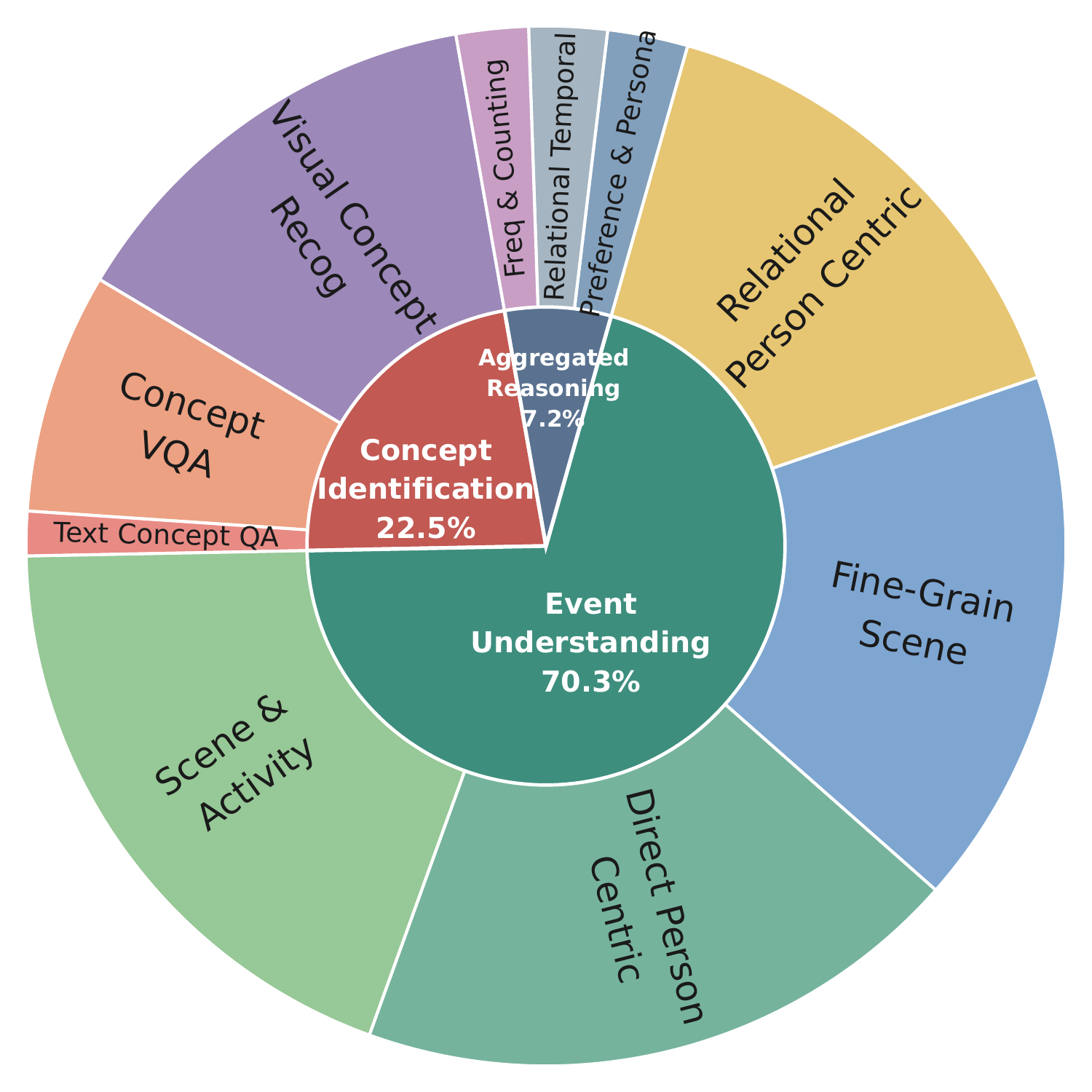}
    \caption{}
    \label{fig:task_pie}
  \end{subfigure}
  \hfill
  \begin{subfigure}[b]{0.73\linewidth}
    \centering
    \includegraphics[width=\linewidth]{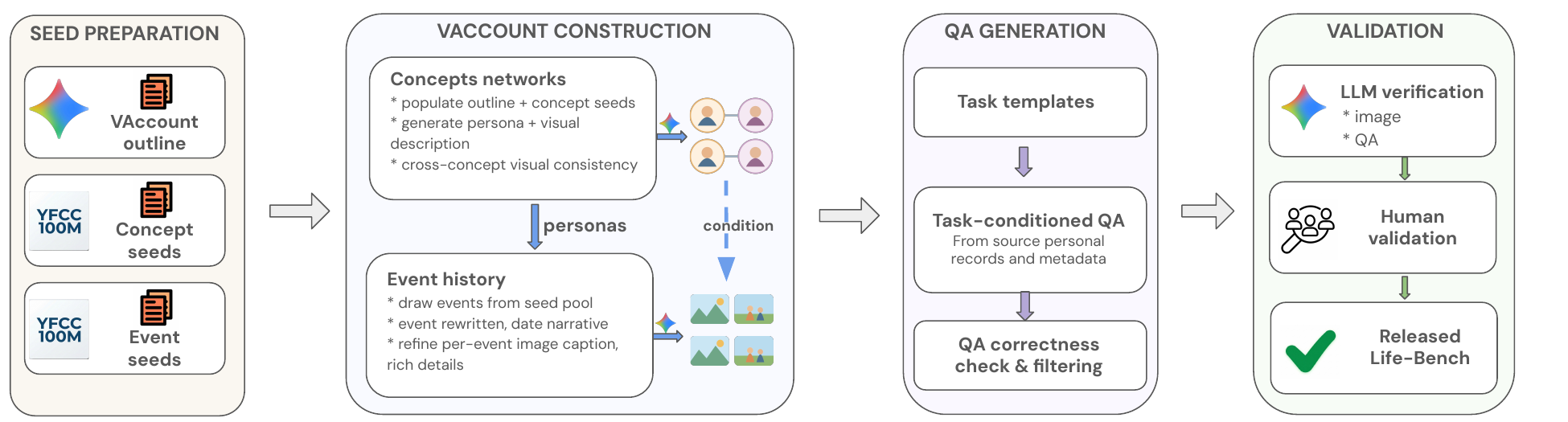}
    \caption{}
    \label{fig:vaccount_construction}
  \end{subfigure}
  \caption{Overview of {\ourdataset}. (a) Distribution of questions across tasks and three categories. (b) The construction pipeline\nop{: synthetic account outlines and textual seeds derived from real-world photo collections build concept networks and multimodal event histories, retained as source records for task-conditioned QA generation, followed by automatic verification and full-coverage human validation before release}.}
  \label{fig:data_overview}
\end{figure*}

We introduce {\ourdataset}, a comprehensive benchmark for evaluating advanced personalization over personal multimodal histories. 
{\ourdataset} comprises over 11.8k human-verified question-answer pairs 
spanning 10 tasks, organized into three categories that probe progressively deeper personal understanding: identifying who is in a user's life~(\textit{Concept Identification}), understanding what happened in the user's history~(\textit{Event Understanding}), and synthesizing what these experiences reveal about the user~(\textit{Aggregated Reasoning}). 
All data is fully synthetic to fundamentally preserve privacy, while the generation process is seeded with textual descriptions generated from real-world photos to retain realism. 

\subsection{Benchmark Design Rationale}
\label{subsec:design_rationale}


Two challenges shape Life-Bench's design. 
The first challenge is obtaining data. Building the benchmark from real individuals would require releasing their photos, social relationships, and long-term records, which privacy and legal concerns make impractical for a public benchmark.
We instead generate {\ourdataset} through a controlled synthetic pipeline seeded with textual scenarios from real-world photo captions.
The generated accounts map to no real individual, yet the seeds retain event scenarios from real photo collections; 
\cameready{in Sec~\ref{subsec:dataset_validation}, we check their scenario diversity against real accounts in embedding space.}

The second challenge lies in task design, motivated by the personalization requirements of real-world applications,
we organize these tasks based on the scope of personal context required by a query.
This yields three levels of context scope: who is in the user's life (personal concepts and social relations); what happened in their history (specific events); and what patterns emerge across it (preferences, activity frequencies, and temporal relations).
Answering them requires context ranging from a single concept, to a specific event, to evidence aggregated across the whole history.
Accordingly, {\ourdataset} groups its tasks into three categories. 

\subsection{Task Categories and Definitions}
\label{subsec:task_categories}
{\ourdataset} comprises 10 tasks.
Questions span two input formats~(text-only, text+image) and three output formats~(multiple-choice, binary, open-generation).
\cref{fig:bench_example} provides an overview and one representative example per task.
The three categories follow their required scope of personal context:
concept-level context, specific events, and evidence distributed across the user's history.
Many tasks require 
cross-modal evidence chains that resolve textual references and connect them
to identities, attributes, actions, or scene details observed in images.
\cref{fig:task_pie} shows the distribution of questions across tasks and categories.
We define each category below \notincamera{and provide task-level details in the Appendix}.




\runin{Concept Identification.}
This category evaluates the foundational ability to recognize and reason about personal concepts: the people and pets in a user's life, referred to by name or relation.
Questions range from direct queries about a single concept to multi-hop queries that require resolving social relation chains.
For image-conditioned questions, the resolved concept must then be linked to its visual appearance to determine its presence, attributes, or actions.
For example, determining whether ``aunt's dog'' appears in a query image requires resolving the referent through the relation chain, followed by cross-modal identity matching against the animal in image.
This category comprises three tasks: \textit{Text Concept QA}, \textit{Visual Concept Recognition}, and \textit{Concept VQA}.

\runin{Event Understanding.}
This category evaluates the ability to locate and reason over specific events in a user's multimodal history.
An event-level answer 
requires temporal or semantic cues to identify the event, visual evidence for fine-grained event details, and relational associations to map observed individuals back to the user's personal context.
These dependencies require cross-modal information integration.
For example, answering what Zosime's son wore on 2013-01-12 requires
identifying David through the relation, locating the dated event, and examining its image.
This category comprises four tasks: \textit{Scene and Activity}, \textit{Direct Person-Centric}, \textit{Relational Person-Centric}, and \textit{Fine-Grained Scene}.

\runin{Aggregated Reasoning.}
This category evaluates the ability to aggregate evidence across multiple dates and events to infer personal patterns, activity frequencies, and relational-temporal sequences.
We evaluate aggregated reasoning in three tasks:
\textit{Preference and Persona} synthesizes recurring evidence across events to infer stable preferences or traits; \textit{Frequency and Counting} requires exhaustive identification of events that satisfy the query conditions; and \textit{Relational Temporal Reasoning} identifies an anchor event and resolves its relation to preceding or subsequent events.
Because individual events may themselves require cross-modal resolution of
descriptions, identities, and activities, this category combines event-level multimodal understanding with history-level aggregation.

\subsection{Benchmark Construction}
\label{subsec:dataset_constru}

{\ourdataset} organizes each synthetic personal history as a \textbf{Vaccount}, the basic unit of personal multimodal context.
Each Vaccount consists of a social network of 3 to 5 concepts including a primary concept~(account owner), and a multimodal history of timestamped images and descriptions.
\cref{fig:vaccount_construction} summarizes the construction pipeline, which prepares textual seeds, constructs Vaccount, and derives task-conditioned question-answer~(QA) pairs from resulting records.
We use Gemini 2.5 Pro~\cite{google25gemini25} for text generation and Gemini 2.5 Flash Image~\cite{gemini25image} for image generation.

\runin{Seed Preparation.}
Three types of textual seeds separately control the account structure and the real-world scenario content of each Vaccount.
An LLM generated synthetic \textit{Vaccount seed} defines the account-level outline: a primary concept with name, age, and career; a social network of 3 to 5 concepts; and a brief life story.
The \textit{concept seeds} and \textit{event seeds} supply realistic subjects and event scenarios to populate this outline.
To obtain these seeds, the LLM generates descriptive captions for images from a CC-licensed subset of the YFCC100M photo dataset~\cite{thomee2016yfcc100m}.
Captions centered on a single person or animal form the concept-seed pool.
The remaining captions are grouped by YFCC account and date, and the LLM synthesizes each group into a narrative event \nop{description that forms an event} seed.
YFCC source images~\cite{thomee2016yfcc100m} serve only as captioning inputs and never enter Vaccounts; all {\ourdataset} images are generated anew to preserve privacy.

\runin{Vaccount Construction.}
This stage builds a coherent concept network and multimodal event history from the account outline.
Given a Vaccount seed, the LLM populates its social network by selecting a concept seed for each listed person or animal.
It then combines each selected seed with the account outline to generate the corresponding concept’s persona and visual description.
These outputs specify each concept's identity, background, and appearance within the Vaccount.
The visual descriptions are jointly refined for logical and cross-concept visual consistency~(e.g., shared facial features among family members). 
The image generation model then converts each refined visual description into a concept portrait, which serves as the identity reference for subsequent history generation.

To build a history consistent with these concepts, the LLM selects event seeds and rewrites each around the established personas as a coherent event narrative with a synthetic date.
For each event, the LLM refines its associated captions into image-level descriptions that depict individual scenes and explicitly name the participating concepts.
The image generation model synthesizes each historical image from its caption, conditioned on the involved concepts' portraits to maintain their visual identities across events.
The resulting Vaccount contains structured records of concept, relation, and a timestamped multimodal history in which images from the same event share the same date and title.

\runin{Question-Answer Generation.}
QA pairs are generated using task-specific templates, task-relevant Vaccount records, and their retained source information.
Vaccount construction retains the source information underlying these records: concept identities, concept relations, personas, event dates, event narratives, image captions, participating concepts, and links between records and generated images.
For each task, the LLM follows the template to generate a question–answer pair from the corresponding Vaccount records, including images, using their source information to ground the reference answer.
Each QA pair is thereby linked to the specific records and images supporting it, enabling the image-fidelity and QA-correctness checks in \cref{subsec:dataset_validation}.

The resulting 11.8k+ QA pairs span 10 Vaccount contexts; \cref{tab:dataset_statistics} summarizes their concept networks, multimodal histories, and question distributions.

\subsection{Quality Control and Benchmark Validity}
\label{subsec:dataset_validation}


\runin{Multi-stage LLM Verification.} This verification targets
intra Vaccount inconsistency, visual deviations from intended facts, and QA pairs unsupported by personal context.
During Vaccount construction, the LLM jointly revises concept profiles and event narratives to keep attributes, relations, and event details consistent with the user profile.
During image generation, we produce 3 candidates per concept portrait or historical image and retain 1 by majority vote among LLM verifiers.
The retained images then undergo multi-round checks for caption fidelity, accurate concept depiction, and factual or visual anomalies, preserving the visual facts behind the QA groundtruth.
During QA generation, the verifier jointly assesses each question and reference answer against its gold context, retaining only unambiguous, answerable, and explicitly supported pairs.
Finally, it evaluates each question without personal context, removing those 
it both deems answerable and answers correctly, thereby filtering items solvable by generic logic or visual shortcuts.

\begin{table}[t]
\centering
\small
\caption{Full-coverage human validation statistics of Life-Bench. All items marked Not Sure and Reject are re-examined and corrected by authors. \nop{Every generated image and QA pair is individually reviewed. Accept, not-sure, and reject rates are reported over all reviewed items; reject reasons are reported over rejected items within each dimension.}}
\label{tab:human_validation}
\resizebox{0.8\linewidth}{!}{%
\begin{tabular}{lccc}
\toprule
& \textbf{Accept} & \textbf{Not Sure} & \textbf{Reject} \\
\midrule
\textbf{Image Fidelity} & 88.20\% & --- & 11.80\% \\
\addlinespace[1pt]
\multicolumn{4}{l}{\textit{Reject reasons~(\% of rejected images)}} \\
\quad Face/Body Distortion        & & & 41.00\% \\
\quad Artifacts / Glitches        & & & 32.40\% \\
\quad Other Visual Flaws          & & & 26.60\% \\
\midrule
\textbf{QA Correctness} & 95.55\% & 2.35\% & 2.10\% \\
\addlinespace[1pt]
\multicolumn{4}{l}{\textit{Reject reasons~(\% of rejected QA pairs)}} \\
\quad Incorrect Answer            & & & 75.09\% \\
\quad Image Inconsistency         & & & 15.79\% \\
\quad Unanswerable / Other        & & & 9.12\% \\
\bottomrule
\end{tabular}%
}
\end{table}

\begin{table}[t]
  \centering
  \caption{\cameready{Caption-embedding alignment and in-account diversity of {\ourdataset} Vaccounts \vs real YFCC accounts.}}
  \label{tab:realism}
\resizebox{0.8\linewidth}{!}{%
  \begin{tabular}{lcc}
    \toprule
    Metric & Vaccount & Real (YFCC) \\
    \midrule
    FID(100)       & 0.56 & 0.53 \\
    Avg.\ intra-account Dist  & 0.91 & 0.96 \\
    \bottomrule
  \end{tabular}%
  }
\end{table}

\runin{Full-Coverage Human Validation.}
Over 100 independent annotators collectively review every image and QA pair before release.
Each image receives an accept/reject judgment for face or body distortion, generation artifacts, and other defects, with the rejection reason recorded.
Given the question, reference answer, and full gold context, annotators label each QA pair as accept, abstain, or reject based on answer correctness and support.
Abstention (2.35\%) preserves uncertain judgments for adjudication.
\Cref{tab:human_validation} reports these initial judgments and failure breakdowns before remediation.
The authors adjudicate every rejected or abstained item: rejected images are regenerated, re-screened by LLM verification, and author-reviewed;
incorrect answers are replaced with confirmed annotator corrections; ambiguous or unanswerable questions are revised.
No item enters the benchmark on an uncertain judgment: each is either confidently accepted by an annotator or explicitly adjudicated by the authors.
\Cref{tab:human_validation} therefore reports errors caught and corrected by full-coverage validation, not residual errors in the released data.

\runin{Distributional Alignment to Real Data and Task Validity.}
We assess whether Vaccounts match the account-level distribution and within-account diversity of real YFCC histories, and whether their questions are solvable by direct semantic matching.
All three analyses use joint caption–metadata embeddings from Gecko~\cite{lee2024gecko}, which also powers our retrieval baselines in \cref{sec:exp}.
For distributional alignment, the Fr\'echet distance~\cite{heusel2018gans} between 10 Vaccounts and 100 randomly sampled YFCC seed accounts is 0.56, close to the real-to-real reference of 0.53 (\cref{tab:realism}).
For within-account diversity, the average pairwise cosine distance is 0.91 for Vaccounts versus 0.96 for real accounts, indicating comparable 
variance. 
For task validity, questions are typically 0.07 closer 
in embedding space to a non-gold context than to the gold context, 
indicating that direct semantic matching does not reliably identify the gold context.
\cameready{These embedding-level checks serve as a sanity check on scenario diversity rather than validating higher-level social or temporal realism.}

\section{{\lifegraph}  for Personalization}
\label{sec:algo}

\begin{algorithm}[t]
    \caption{{\lifegraph} Retriever $\mathcal{R}_G$}
    \small
    \label{alg:retrieval}
    \begin{algorithmic}[1]
    \renewcommand{\algorithmicrequire}{\textbf{Require:}}
    \renewcommand{\algorithmicensure}{\textbf{Ensure:}}

    \REQUIRE Input query $x$, VLM $\mathcal{F}$, a {\lifegraph} $G=(N,E,T)$ with source links $\mathrm{src}(\cdot)$, depth limit $d$, path width $k$
    \ENSURE Retrieved context $C_x = C_{paths} \cup C_{refs}$

    \STATE \textbf{// Initialization}
    \STATE $\mathbf{n}^0 \leftarrow \textsc{TopEntities}(x, G, k)$
    \STATE $C_{paths} \leftarrow \{\mathbf{n}^0\}$; $C_{refs} \leftarrow \emptyset$

    \FOR{$\bar{d} = 1$ \textbf{to} $d$}
        \STATE \textbf{// Search}: \textit{$C_{path}$ collect unvisited relation–entity neighbors of $\mathbf{n}^{\bar{d}-1}$}
        \STATE $(\mathbf{e}_{cand}^{\bar{d}}, \mathbf{n}_{cand}^{\bar{d}}) \leftarrow \textsc{Search}(G, \mathbf{n}^{\bar{d}-1}, C_{paths})$

        \STATE \textbf{// Pruning}: \textit{Select the top-$k$ ``relation--entity'' pairs most relevant to $x$}
        \STATE $(\mathbf{e}^{\bar{d}}, \mathbf{n}^{\bar{d}}) \leftarrow \textsc{Prune}(\mathcal{F}, x, C_{paths}, k, (\mathbf{e}_{cand}^{\bar{d}}, \mathbf{n}_{cand}^{\bar{d}}))$

        \STATE \textbf{// Path Update}
        \STATE $C_{paths} \leftarrow C_{paths} \cup \{(\mathbf{e}^{\bar{d}}, \mathbf{n}^{\bar{d}})\}$

        \STATE \textbf{// Source Reference:} \textit{Fetch $\mathrm{src}(g)$ for selected $g \in C_{paths}$ that require source evidence}
        \STATE $C_{refs} \leftarrow \textsc{FetchReference}(\mathcal{F}, x, C_{paths})$

        \STATE \textbf{// Sufficiency Check:} \textit{Terminate once retrieved context suffices to answer $x$}
        \IF{$\textsc{Reasoning}(\mathcal{F}, x, C_{paths}, C_{refs})$}
            \STATE \textbf{break}
        \ENDIF
    \ENDFOR
    \STATE $C_x \leftarrow C_{paths} \cup C_{refs}$
    \STATE \textbf{return} $C_x$
    \end{algorithmic}
\end{algorithm}

\subsection{Problem Formulation}
\label{subsec:problem_definition}

With multimodal models increasingly serving as personal agents, personalized queries require drawing on a user's growing multimodal history, motivating retrieval of relevant evidence instead of repeatedly providing the complete history.
Let $\mathcal{F}$ be a VLM, $P$ a user's multimodal history, and $\mathcal{R}$ a retriever; for a query $x$, retrieval-enhanced personalization is formalized as:
\begin{equation}
  y = \mathcal{F}(x, C_x) \quad \text{where} \quad C_x = \mathcal{R}(x, P)
  \label{eq:def_personalization}
\end{equation}

Personalization over multimodal histories requires aligning evidence across text and visual records such as linking a person named to their visual appearance, while reasoning over relationships and events distributed throughout the history.
When text-based retrieval represents images as captions, this textualization omits fine-grained visual evidence such as appearance, actions, spatial relations.
Without organizing this information in advance, query-time retrieval must reconstruct cross-modal personal context and locate original visual evidence  omitted by textual representations.

These requirements motivate LifeGraph, an end-to-end personal knowledge graph framework that organizes concepts, relations, events, and temporal connections over multimodal histories.
At construction time, LifeGraph extracts identities, relations, and events from multimodal histories and encodes them as structured knowledge, rather than merely indexing source data.
The graph also retains provenance links to the source multimodal records, allowing retrieval to combine stored knowledge with multimodal evidence when fine-grained visual details are needed.

\subsection{{\lifegraph} Construction}
\label{subsec:lifegraph}

Given a user's personal context $P$, 
comprising concept and multimodal history records,
{\lifegraph} uses a VLM to extract personal concepts, relations, and facts from $P$ \nop{these records} and organize them as a directed personal knowledge graph $G$\nop{before retrieval}.
Formally, $G=(N,E,T)$, where $N$ denotes the entity nodes, $E$ the relation vocabulary, and $T \subseteq N \times E \times N$ the directed relational triples; each $(u,r,v) \in T$ encodes a personal fact.
{\lifegraph} uses a hybrid schema that fixes node types while leaving relations open-ended:
entity nodes are assigned one of six predefined types (PersonAnimal, Event, Date, Location, Activity, and Object), 
which constrain VLM-based extraction and cross-record normalization, keeping the graph structurally consistent.
Relations may carry auxiliary attributes: to represent complex \textit{n}-ary facts~\cite{luo2024text2nkg,wang2021nary}, {\lifegraph} designates a primary binary relation and records the remaining arguments as edge attributes, yielding a hyper-relational representation within the triple-based graph.

{\lifegraph} constructs $G$ in two stages so that multimodal history records are interpreted using an established representation of the user's personal concepts and their relations.
In the first stage, the VLM builds a concept subgraph $G_c \subseteq G$ from concept records, connecting personal concepts through their relations, and retaining attributes~(\eg name, gender) and links to their images rather than flattening visual details into text descriptions. 
In the second stage, the VLM conditions on $G_c$ while processing batches of multimodal history records, associates mentioned or depicted personal concepts with existing nodes, and extends the graph with event-centered triples connecting events to dates, participants, activities, and other extracted entities.
This grounds depicted individuals and objects in the established concept structure during construction, avoiding their re-identification from standalone captions at query time.
{\lifegraph} represents the implicit connections among historical events through shared concept, date, or object nodes rather than by separately extracting event-event relations.
Following prior VLM-assisted graph construction methods~\cite{gong2024uknow,yang2025eventrag}, both stages use a two-turn extraction protocol to constrain triple generation to an explicit candidate set: 
the first identifies typed candidate entities, and the second generates factual triples among them.
\notincamera{Prompt details are provided in Appendix.}

{\lifegraph} maintains provenance links from extracted entities and facts to their supporting multimodal source records.
During construction, these source references are stored as attributes of the corresponding nodes and triple instances; for any $g \in N \cup T$, $\mathrm{src}(g) \subseteq P$ denotes the records in $P$ linked to $g$.
The resulting graph combines structured personal knowledge with direct links to the original multimodal records, allowing the retriever to navigate the graph and fetch source evidence on demand when a query requires visual details not encoded in the triples.

\subsection{Graph Retrieval with Multimodal Evidence} 
\label{subsec:retrieval}

Given the constructed {\lifegraph}, the graph retriever $\mathcal{R}_G$ retrieves two complementary forms of evidence for a query $x$: relational facts distributed across $G$ and, when $x$ requires visual details absent from those facts, their supporting multimodal records.
Accordingly, $\mathcal{R}_G$ returns $C_x=\{C_{\mathrm{paths}},C_{\mathrm{refs}}\}$, where $C_{\mathrm{paths}}$ contains traversed relational paths and $C_{\mathrm{refs}}$ the source records reached through $\mathrm{src}(\cdot)$.
For graph traversal, $\mathcal{R}_G$ follows the model-guided iterative beam-search paradigm of Think-on-Graph~\cite{sun2024tog}: from query-relevant entry entities, it expands and prunes candidate paths, retaining the top-$k$ paths up to depth $d$.
This traversal follows the established formulation; the design of $\mathcal{R}_G$ instead centers on jointly ranking adjacent relation: entity pairs and retrieving multimodal source evidence through $G$'s provenance links~(\cref{alg:retrieval}).

\cref{alg:retrieval} first applies \textsc{TopEntities} to select the entry nodes $n^0$ most relevant to $x$.
Because the relation-entity candidates at each expansion step can be enumerated together in the per-user {\lifegraph}, \(\mathcal{R}_G\) asks \(\mathcal{F}\) to rank them jointly in one VLM call rather than searching and pruning relations and entities separately~\cite{sun2024tog}.
After pruning, \textsc{Fetch\_Reference} determines whether $x$ requires visual evidence absent from the retrieved paths and, if so, follows $\mathrm{src}(\cdot)$ from their nodes and triple instances to retrieve $C_{\mathrm{refs}}$.
This channel relies on construction-time provenance rather than the structured graph alone.
At the end of each round, $\mathcal{F}$ determines whether the accumulated $C_{\mathrm{paths}}$ and current $C_{\mathrm{refs}}$ suffice to answer $x$, terminating if so and otherwise continuing to depth $d$.
The resulting $C_x$ is passed to $\mathcal{F}$ for response generation as in Eq.~(\ref{eq:def_personalization}). 
Since $C_{\mathrm{paths}}$ is selected from $G$ and $C_{\mathrm{refs}}$ through $\mathrm{src}(\cdot)$, construction and retrieval jointly determine the personal evidence available at inference time.

\section{Experiments}
\label{sec:exp}

\begin{table*}[t]
  \centering
  \caption{Evaluation results across all {\ourdataset} tasks. Retrieval-based methods are compared for context size $k \in \{3,5\}$. Tasks are grouped into 3 categories. The best and second-best scores among each setting are in \textbf{bold} and \underline{underlined} respectively. ($^*$ Oracle overall averages over seven concept and event tasks since aggregate does not have subset golden reference) }
  \label{tab:main_results_merged}
  \resizebox{0.95\textwidth}{!}{%
  \setlength{\tabcolsep}{2pt}
  \footnotesize
  \begin{tabular}{lccccccccccc}
    \toprule
    \multicolumn{1}{c}{\multirow{2}{*}{Method}} & \multicolumn{3}{c}{Concept} & \multicolumn{4}{c}{Event} & \multicolumn{3}{c}{Aggregated} & \multicolumn{1}{c}{\multirow{2}{*}{Overall}} \\
    \cmidrule(lr){2-4} \cmidrule(lr){5-8} \cmidrule(lr){9-11}
    & \makecell{Text Concept\\QA} & \makecell{Visual Concept\\Recognition} & \makecell{Concept\\VQA} & \makecell{Scene \&\\Activity} & \makecell{Direct\\Person-Centric} & \makecell{Relational\\Person-Centric} & \makecell{Fine-Grained\\Scene} & \makecell{Preference\\~\& Persona} & \makecell{Relational\\Temporal} & \makecell{Frequency\\~\& Counting} & \\
    \midrule
    Random & 0.2500 & 0.5000 & 0.2500 & 0.0000 & 0.0000 & 0.0000 & 0.0000 & 0.2003 & 0.0000 & 0.0000 & 0.1200 \\
    Close-book & 0.2454 & 0.5111 & 0.2696 & 0.0848 & 0.0539 & 0.1341 & 0.1560 & 0.4373 & 0.0662 & 0.0755 & 0.2034 \\
    Full-context & 0.9730 & 0.9722 & 0.9508 & 0.9143 & 0.9095 & 0.9141 & 0.9014 & 0.8856 & 0.8523 & 0.3823 & 0.8655 \\
    Oracle & 0.9816 & 0.9826 & 0.9570 & 0.9324 & 0.9292 & 0.9269 & 0.9100 & -- & -- & -- & 0.9457$^*$ \\
    Oracle-Gemma & 0.8527 & 0.8698 & 0.6170 & 0.8093 & 0.7007 & 0.7130 & 0.7410 & -- & -- & -- & 0.7576$^*$ \\
    \midrule
    \midrule
    BM25$_{k=3}$~\cite{robertson2009bm25} & \underline{0.8160} & 0.8177 & \underline{0.5740} & 0.6155 & 0.3417 & 0.4409 & 0.4740 & 0.4780 & 0.1185 & 0.1359 & 0.4812 \\
    RAG-Cap$_{k=3}$~\cite{lewis2020rag} & \textbf{0.8221} & 0.7436 & 0.5135 & 0.4371 & 0.2517 & 0.3377 & \textbf{0.5257} & 0.6373 & 0.1254 & 0.1396 & 0.4534 \\
    RAP$_{k=3}$~\cite{hao2025rap} & 0.4513 & 0.3976 & 0.4478 & 0.3316 & 0.2227 & 0.3350 & 0.3074 & 0.5525 & 0.1603 & 0.0943 & 0.3301 \\
    RAP-Gemma$_{k=3}$~\cite{hao2025rap} & 0.7607 & 0.7460 & 0.4896 & 0.2742 & 0.1252 & 0.2086 & 0.4435 & \textbf{0.6712} & 0.1185 & 0.0943 & 0.3932 \\
    R2P$_{k=3}$~\cite{das2025r2p} & 0.6319 & 0.5439 & 0.3716 & 0.2627 & 0.2005 & 0.2930 & 0.3870 & 0.4678 & 0.1533 & 0.0453 & 0.3357 \\
    R2P-Gemma$_{k=3}$~\cite{das2025r2p} & 0.6430 & 0.6904 & 0.3752 & 0.3068 & 0.2667 & 0.3419 & 0.3956 & 0.5613 & 0.1707 & 0.0679 & 0.3820 \\
    HippoRAG2$_{k=3}$~\cite{gutierrez2025hipporag2} & 0.8159 & 0.8097 & \textbf{0.6090} & 0.3364 & 0.1764 & 0.2274 & 0.3372 & 0.4848 & 0.1429 & 0.1245 & 0.4064 \\
    \hdashline
    {\lifegraph}$_{d=2,k=3}$ & 0.7117 & \textbf{0.8264} & 0.5422 & \textbf{0.7143} & \textbf{0.4575} & \textbf{0.5237} & 0.5081 & \underline{0.6576} & \textbf{0.2927} & \textbf{0.1849} & \textbf{0.5419} \\
    {\lifegraph}$_{d=3,k=3}$ & 0.7791 & \underline{0.8262} & 0.5364 & \underline{0.7108} & \underline{0.4486} & \underline{0.5072} & \underline{0.5106} & 0.6475 & \underline{0.2822} & \underline{0.1509} & \underline{0.5400} \\
    \midrule
    \midrule
    BM25$_{k=5}$~\cite{robertson2009bm25} & \textbf{0.8650} & 0.8320 & \underline{0.6249} & 0.6698 & 0.3648 & 0.4923 & 0.4962 & 0.5593 & 0.1080 & 0.1472 & 0.5159 \\
    RAG-Cap$_{k=5}$~\cite{lewis2020rag} & \underline{0.8405} & 0.7650 & 0.5236 & 0.4843 & 0.2895 & 0.3786 & \textbf{0.5671} & \underline{0.7051} & 0.1428 & 0.1623 & 0.4859 \\
    RAP-Gemma$_{k=5}$~\cite{hao2025rap} & 0.7651 & 0.7498 & 0.4744 & 0.3043 & 0.1688 & 0.2467 & 0.5172 & \textbf{0.7119} & 0.1533 & 0.0755 & 0.4167 \\
    R2P-Gemma$_{k=5}$~\cite{das2025r2p} & 0.7055 & 0.6943 & 0.3559 & 0.3244 & 0.2664 & 0.3488 & 0.4338 & 0.5831 & 0.1715 & 0.1094 & 0.3993 \\
    HippoRAG2$_{k=5}$~\cite{gutierrez2025hipporag2} & 0.8098 & \textbf{0.8494} & \textbf{0.6542} & 0.3978 & 0.2094 & 0.3201 & 0.4356 & 0.5559 & 0.1672 & 0.1623 & 0.4562 \\
    \hdashline
    {\lifegraph}$_{d=2, k=5}$ & 0.7587 & \underline{0.8332} & 0.5318 & \textbf{0.7242} & \textbf{0.5092} & \underline{0.5193} & 0.5157 & 0.6746 & \underline{0.3240} & \underline{0.1785} & \underline{0.5569} \\
    {\lifegraph}$_{d=3, k=5}$ & 0.7914 & 0.8278 & 0.5260 & \underline{0.7228} & \underline{0.4984} & \textbf{0.5348} & \underline{0.5253} & 0.6524 & \textbf{0.3275} & \textbf{0.1976} & \textbf{0.5604} \\
    \bottomrule
  \end{tabular}%
  }
\end{table*}

\begin{figure*}
    \centering
    \includegraphics[width=0.9\linewidth]{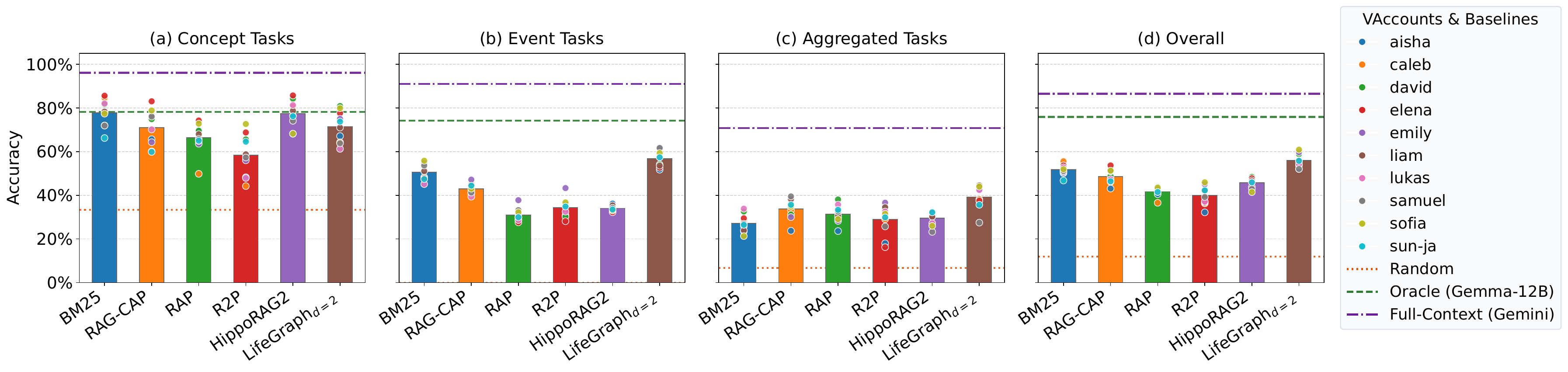}
    \caption{Per-Vaccount accuracy of retrieval methods at $k=5$ (LifeGraph: $d=2$), grouped by task category. Bars show the mean across 10 Vaccounts, dots show individual Vaccounts.}
    \label{fig:vaccount_category_hist}
\end{figure*}

\subsection{Experimental Setup}
\label{subsec:exp_set}

We evaluate {\lifegraph} and representative retrieval baselines across all 10 tasks in {\ourdataset}, together with reference settings that contextualize the reported performance.

\noindent\textbf{Baselines.}
All retrieval methods follow the retrieve-then-generate interface~(Eq.~\ref{eq:def_personalization}), differing primarily in how they organize and retrieve personal context.
The baselines cover four retrieval paradigms:
(1) BM25~\cite{robertson2009bm25}: sparse lexical retrieval that ranks textual records and image captions by term matching;
(2) RAG-Cap~\cite{lewis2020rag}: dense retrieval that performs embedding retrieval~(Gecko-1B~\cite{lee2024gecko}) over descriptive image captions;
(3) RAP~\cite{hao2025rap} and R2P~\cite{das2025r2p}, two multimodal personalization methods that retrieve from key-value concept memory and distinctive concept ``fingerprints'' respectively;
(4) HippoRAG2~\cite{gutierrez2025hipporag2}, structured text retrieval that constructs a graph indexing and retrieves evidence using Personalized PageRank.
BM25, RAG-Cap, and HippoRAG2 index the same caption corpus, allowing a controlled comparison of lexical, dense, and graph-based retrieval over the same textual representation.
All methods deliver retrieved records to the reasoning backbone in the same image+text form.

\noindent\textbf{Diagnostic Baselines.}
We further report 4 reference settings  to contextualize retrieval performance, all generated with Gemini-3.5-Flash~\cite{gemini35flash2026} except random.
(1) Random reports mathematical randomization performance, (2) Close-book answers without personal context, showing what model priors alone can solve.
(3) Oracle receives only the gold supporting evidence, isolating generation and reasoning errors under perfect retrieval.  
We also report it with the standardized Gemma backbone. 
(4) Full-context places an entire Vaccount history in the prompt; we treat it only as a diagnostic reference, not practical solution: real personal archives far exceed model context windows, hundreds of images per query requires tens of times more tokens than retrieving a few relevant records.


\noindent\textbf{Evaluation Metrics.}
Our primary metric is accuracy: multiple-choice and binary questions use exact match, 
while Gemini 3.6 Flash~\cite{gemini36flash2026} judges open-ended responses against ground-truth answer using LLM-as-a-Judge~\cite{zheng2023judging}. 
We additionally report recall to assess retrieval quality.
To validate this automatic evaluation, we uniformly sample 2,090 instances covering every Vaccount and task over five representative settings, and rescore them with GPT-5.4~\cite{gpt54} as an independent judge from another model family.
The two judges agree on 96.2\% of samples (Cohen's $\kappa = 0.92$~\cite{cohen1960coefficient}), with full method ranking unchanged.
GPT-5.4 is generally stricter but 
provides no evidence that Gemini judge substantially favors responses from its own family in this sample.
Moreover, the gap between {\lifegraph} and the strongest baseline widens 
under GPT-5.4, suggesting that our reported improvements are conservative. 
\notincamera{Full per-task agreement scores are detailed in Appendix.}


\noindent\textbf{Implementation Details.} 
For fair comparison, all retrieval methods use Gemma-3 12B~\cite{team2025gemma} as backbone, better isolating differences among retrieval frameworks.
Retrieval baselines are evaluated with context size $k \in \{3, 5\}$;
{\lifegraph} uses the same $k$ as retrieve width, with maximum path depth $d \in \{2, 3\}$.
For completeness, we also report RAP~\cite{hao2025rap} and R2P~\cite{das2025r2p} with their originally proposed backbones~(LLaVA-13B~\cite{liu2024llava15}, MiniCPM-o~\cite{yao2024minicpm} respectively); due to their limited context windows, these results are reported only for $k=3$.
We denote these configurations as Method$_{k|d}$.

\subsection{Main Findings}
\label{subsec:main_results}

We organize our main findings around the end-to-end accuracy results in \cref{tab:main_results_merged}, the recall reference in \cref{tab:recall}, and the per-Vaccount category-wise analysis in \cref{fig:vaccount_category_hist}.

\noindent\textbf{Performance degrades as required evidence scope and reasoning complexity increases.}
As shown in \cref{tab:main_results_merged} and \cref{fig:vaccount_category_hist}, average accuracy across retrieval methods ranges from $0.58$-$0.77$ on Concept Identification tasks, but decreases to $0.30$-$0.57$ on Event Understanding and $0.26$--$0.40$ on Aggregated Reasoning. 
In particular Aggregated tasks which require relation resolution together with temporal navigation or exhaustive aggregation over the full history: the best retrieval results reach only $0.33$ on Relational Temporal Reasoning and $0.20$ on Frequency and Counting.
Even full-context, which bypasses retrieval, achieves only $0.38$ on Frequency and Counting.
This performance gradient follows the intended evidence-scope design of {\ourdataset}, providing empirical support for its task structure and showing that history-wide aggregation remains an open challenge for current systems.

\noindent\textbf{Both retrieval and reasoning capabilities constrain performance.}
Close-book achieves only $0.20$ overall, compared with $0.12$ for random guessing, showing that most questions cannot be solved without personal context.
Retrieval constitutes a major bottleneck: under the same Gemma backbone, macro-averaged accuracy over the 4 Event tasks decreases from $0.74$ with oracle evidence to $0.57$ with the strongest retrieval configuration, leaving a gap of $0.17$.
Moreover, no method exceeds $0.40$ average accuracy on Aggregated tasks~(\cref{fig:vaccount_category_hist}); 
consistently, no method exceeds $0.56$ overall Recall@5 in \cref{tab:recall}, leaving nearly half of the gold source events unretrieved.
Beyond retrieval, reasoning capacity remains important even when relevant evidence is provided. 
On the seven tasks with oracle evaluation, Oracle Gemini-3.5-Flash reaches $0.95$ overall, whereas Oracle Gemma-3-12B reaches $0.76$,
revealing a $0.19$ backbone-dependent generation and reasoning gap even after retrieval errors are removed.

\noindent\textbf{No single retrieval paradigm dominates: performance depends on the alignment between the demanded evidence scope and the method's retrieval and storage strategy.}
Table~\ref{tab:main_results_merged} covers four paradigms: lexical (BM25~\cite{robertson2009bm25}), dense (RAG-Cap~\cite{lewis2020rag}), concept-memory (RAP~\cite{hao2025rap}, R2P~\cite{das2025r2p}), and graph (HippoRAG2~\cite{gutierrez2025hipporag2}, {\lifegraph}), whose ranking reverses from task to task.
Most tellingly, RAP and R2P, the two methods built for multimodal personalization, rank last overall (0.417 and 0.399, $k=5$), consistent with their low recall (R@5 0.461 and 0.427).
This reflects what they index: concept and persona-level memory without event-level history. 
While RAP accordingly leads Preference \& Persona (0.712),  both collapse on event questions (Direct Person-Centric at most 0.266 \vs 0.365 for BM25).
Over the shared caption corpus, BM25 is the stronger baseline overall (0.516 \vs 0.486). This is expected as personal context is marked by distinctive keywords (names, relations, dates) that lexical matching directly captures.
It also achieves the top in Text Concept QA (0.865).
RAG-Cap instead wins Fine-Grained Scene (0.567 \vs 0.496), where answers hinge on paraphrased scene descriptions that embeddings match and keywords miss.
HippoRAG2 posts the best visual-concept scores of any method (Visual Concept Recognition 0.849, Concept VQA 0.654) through multi-hop retrieval over its entity graph, but does not maintain this advantage on event and aggregated tasks: its graph links general entities rather than personal specific historical events, and offers no path back to visual sources.
{\lifegraph} leads overall, with the largest margins on event tasks (Scene \& Activity 0.723, Relational Person-Centric 0.535) and aggregated tasks (Relational Temporal 0.328, Frequency \& Counting 0.198). Because these tasks require locating events across modalities and reasoning over them jointly, they align perfectly with the structure of {\lifegraph}'s personal, event-centric, source-linked graph. Conversely, it trails on concept tasks, where the required evidence is concept-level and event-centric organization confers no additional advantage.
What differentiates these methods apart is not their underlying retrieval machinery, but rather the unit each approach indexes~(granularity, modality, and knowledge organization), the exact dimension {\ourdataset} is built to measure.

\noindent\textbf{Increasing retrieval width yields diminishing gain.}
From $k=3$ to $k=5$, overall recall increases from $0.39$ to $0.47$ for RAG-Cap and from $0.36$ to $0.45$ for HippoRAG2, while their overall accuracy improves by only $3.3$ and $5$ points, respectively.
At the extreme, full-context Gemini-3.5-Flash, which is supplied with the entire history, only achieves $0.382$ on Frequency and Counting. This is consistent with recent studies that frontier models struggle with long-context aggregation and counting~\cite{bertsch2025oolong}.
These results indicate that simply increasing the retrieval budget cannot substitute for precise localization and organization of personal evidence.

\begin{table}[t]
\caption{Event-level Recall@3/5 on {\ourdataset}. Retrieving any image from target event counts as a hit. Recall is used as an auxiliary retrieval diagnostic rather than directly comparable method ranking, since {\lifegraph} may answer from graph-encoded facts without retrieving source records.}
\label{tab:recall}
\centering
\small
\resizebox{\columnwidth}{!}{%
\begin{tabular}{l cc cc cc}
\toprule
\multirow{2}{*}{Method} & \multicolumn{2}{c}{Concept} & \multicolumn{2}{c}{Event} & \multicolumn{2}{c}{Overall} \\
\cmidrule(lr){2-3} \cmidrule(lr){4-5} \cmidrule(lr){6-7}
 & R@3 & R@5 & R@3 & R@5 & R@3 & R@5 \\
\midrule
BM25~\cite{robertson2009bm25}             & 0.4626 & \textbf{0.5330} & 0.4785 & 0.5371 & 0.4717 & 0.5353 \\
RAG-Cap~\cite{lewis2020rag}          & 0.4307 & 0.4808 & 0.3665 & 0.4580 & 0.3940 & 0.4678 \\
R2P~\cite{das2025r2p}        & 0.4070 & 0.4238 & 0.3823 & 0.4299 & 0.3929 & 0.4273 \\
RAP~\cite{hao2025rap}        & 0.4354 & 0.4653 & 0.4065 & 0.4572 & 0.4188 & 0.4607 \\
HippoRAG2~\cite{gutierrez2025hipporag2}        & 0.4465 & \underline{0.5128} & 0.3042 & 0.4065 & 0.3652 & 0.4521 \\
\midrule
$\text{LifeGraph}_{d=2}$  & \underline{0.4748} & 0.5009 & \textbf{0.5309} & \textbf{0.6021} & \underline{0.5068} & \textbf{0.5587} \\
$\text{LifeGraph}_{d=3}$  & \textbf{0.4844} & 0.4991 & \underline{0.5271} & \underline{0.5963} & \textbf{0.5088} & \underline{0.5546} \\
\bottomrule
\end{tabular}%
}
\end{table}

Findings above remain consistent at Vaccount-level. 
Figure~\ref{fig:vaccount_category_hist} breaks down each method's category-wise accuracy by Vaccount, showing the method ordering holds on all 10 Vaccounts, and per-account spread is visibly smaller than the gaps between methods.

\subsection{{\lifegraph} Analysis}
\label{subsec:exp_ablation}



\begin{figure}[t]
  \centering
  \includegraphics[width=0.8\linewidth]{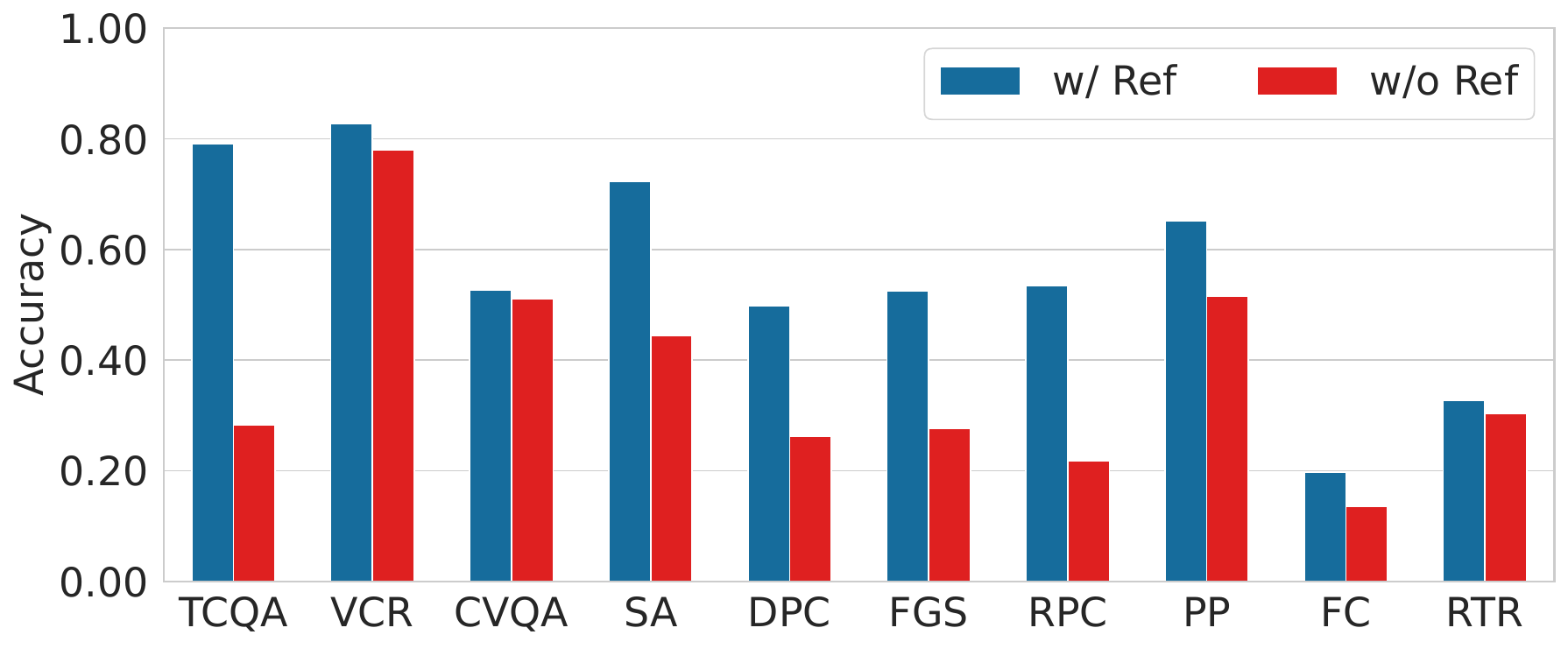}
  \caption{Accuracy of {\lifegraph}$_{d=3, k=5}$ with (w/ Ref) and without (w/o Ref) multimodal source retrieval across tasks. Tasks are denoted by its acronym.
  \nop{Impact of source data retrieval for {\lifegraph}. The bars show the change in accuracy for {\lifegraph}$_{d=3, k=5}$ when the multimodal source reference module is disabled~(w/o Ref).} }
  \label{fig:effect_ref_bar}
\end{figure}

\noindent\textbf{Ablation on Source Multimodal Retrieval.}
We analyze the effects of retrieving multimodal source data by ablating it: retrieval traverses the graph as usual, but the model reasons solely on the textual knowledge organized in LifeGraph ($C_{paths}$), without the retrieved source images ($C_{refs}$).
As shown in \cref{fig:effect_ref_bar}, this w/o Ref variant collapses on tasks whose evidence resides in the visual records (\eg~Text Concept QA asks about visual attribution) yet retains strong performance on relational and logical reasoning (\eg~Relational Temporal Reasoning). 
This confirms that the graph and source play complementary roles: the graph structure encodes high-level personal knowledge and maintains relevant evidence, and the source multimodal data supply the fine-grained, visual evidence textualization inevitably discards. 
This design allows {\lifegraph} to cover cross-modality personal understanding. 

\begin{figure}[t]
  \centering
  \includegraphics[width=0.75\linewidth]{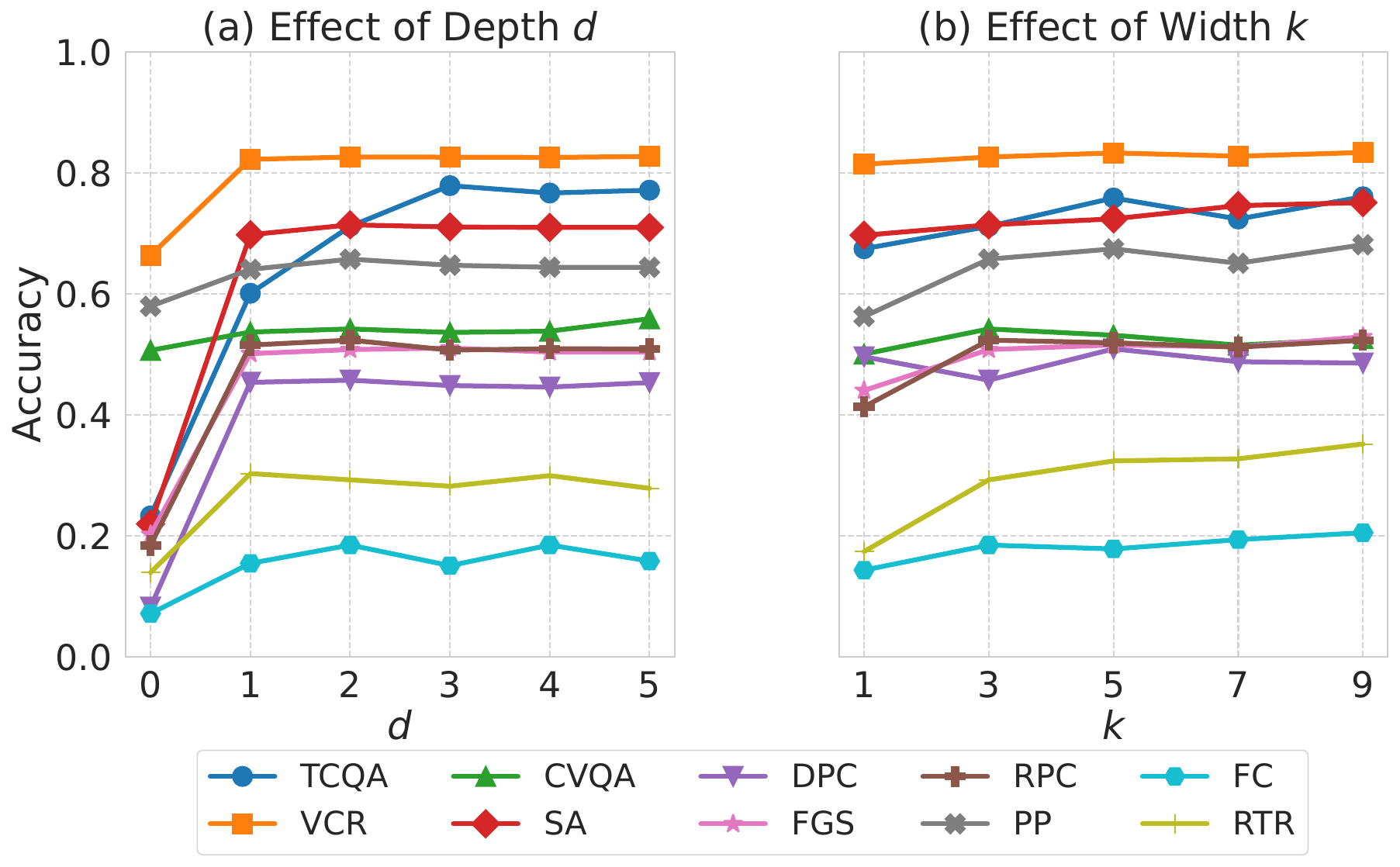}
  \caption{Effect of retrieval depth $d$ ((a), evaluated at $k=3$) and width $k$ ((b), evaluated at $d=2$) on {\lifegraph} performance. Tasks are denoted by its acronym.}
  \label{fig:effect_kd}
\end{figure}

\noindent\textbf{Effects of Retrieval Depth and Width.}
{\lifegraph}
performance saturates under a shallow graph-search budget (\cref{fig:effect_kd}).
Accuracy plateaus by depth $d=2$ for most tasks~(under $k=3$), with concept-retrieval tasks such as Text Concept QA still benefiting from one additional hop ($d=3$).
Since \textit{sufficiency check} terminates retrieval once the accumulated context suffices to answer, the average effective depth stays below 1.9 even when the depth limit is raised to 5\notincamera{~(see Appendix)}.
Width shows a similar trend~(\cref{tab:main_results_merged}, \cref{fig:effect_kd}): 
most tasks plateau at $k$ between 3 and 5 and fluctuate thereafter, while larger $k$ brings modest gains mainly on tasks needing broader event coverage.
This is consistent with the diminishing retrieval gains from $k=3$ to $k=5$ observed across all methods in \cref{subsec:main_results}.

\section{Conclusion}
\label{sec:conclusion}

We introduced {\ourdataset}, a fully synthetic, human-verified multimodal benchmark organizing 10 personalization tasks by required evidence scope, from concept identification to history-wide aggregation. 
Evaluating four retrieval paradigms reveals three findings: accuracy degrades sharply as scope broadens, falling below 0.40 on aggregated tasks; no single paradigm dominates, as each excels only when its index matches the required evidence; and both retrieval and reasoning remain bottlenecks, with substantial headroom even under oracle retrieval.
We also proposed {\lifegraph}, whose structured, source-linked retrieval shows particular promise on event and aggregated tasks, yet performance beyond concept recognition remains modest across all methods, leaving personalization over multimodal histories an open challenge.
Addressing this challenge will demand both richer organization of personal knowledge and stronger cross-modal reasoning over full personal histories.

\section*{Ethical Considerations}


To mitigate the privacy and licensing risks of real personal photo histories, {\ourdataset} is fully synthetic: scenario seeds derive from a CC-licensed YFCC100M subset, with source images used solely for captioning, narratives rewritten across stages, all benchmark images generated anew, and no YFCC images or source-derived captions released. The pipeline and release boundaries underwent internal legal, dataset, and publication reviews, though residual risks such as incidental resemblance to real people or events cannot be ruled out entirely. Full-coverage human validation was conducted through an institutionally managed third-party rater platform under established compliance procedures, with raters compensated on an hourly basis. Because Vaccounts organize individual histories rather than demographic groups and synthetic data may inherit biases from both the source distribution and generative models, {\ourdataset} makes no claim of demographic representativeness and alone should not be used for population-level fairness conclusions.

Both {\ourdataset} and LifeGraph enforce per-account isolation, confining graph construction and retrieval to a single Vaccount without cross-user aggregation or social-graph linkage, thereby limiting cross-user privacy risks. Deploying such systems on real personal archives would require user consent, access controls, mechanisms to inspect, correct, and delete stored facts and source links, and third-party protections for others appearing in a user's archive. Our experiments use only synthetic histories and do not establish readiness for high-stakes deployment.

\bibliography{main}

\appendix
\clearpage
\appendix
\section*{Appendix}

\section{{\ourdataset} Details}

\begin{table*}[t]
\centering
\caption{Comprehensive breakdown of evaluation tasks in \ourdataset.}
\label{tab:suppl_task_breakdown}
\resizebox{\textwidth}{!}{%
\begin{tabular}{@{}lllllr@{}}
\toprule
\textbf{Category} & \textbf{Task} & \textbf{Reasoning Scope / Focus} & \textbf{Input Format} & \textbf{Answer Format} & \textbf{\# Questions} \\
\midrule
\multirow{3}{*}{\textbf{Concept Identification}} 
 & Text Concept QA & Multi-hop Persona \& Visual Attribute Retrieval & Text & Multiple-Choice & 163 \\
 & Visual Concept Recognition & Multi-hop Identity Verification & Text + Image & Binary & 1,613 \\
 & Concept VQA & Multi-hop Grounded Visual Reasoning & Text + Image & Multiple-Choice & 885 \\
\midrule
\multirow{4}{*}{\textbf{Event Understanding}} 
 & Scene and Activity & Holistic Event Context \& Activity Understanding & Text & Open Generation & 2,265 \\
 & Direct Person-Centric & Concept Appearance \& Actions in Events & Text & Open Generation & 2,245 \\
 & Relational Person-Centric & Relational Concept Localization \& Interactions & Text & Open Generation & 1,812 \\
 & Fine-Grained Scene & Compositional Spatial \& Multi-concept Moments & Text & Open Generation & 1,981 \\
\midrule
\multirow{3}{*}{\textbf{Aggregated Reasoning}} 
 & Preference and Persona & Option-guided Preference \& Trait Synthesis & Text & Open Gen (Option-guided) & 295 \\
 & Frequency and Counting & Full-history Quantitative Event Aggregation & Text & Open Generation & 265 \\
 & Relational Temporal Reasoning & Anchor Event Localization \& Temporal Navigation & Text & Open Generation & 287 \\
\midrule
\textbf{Total (Evaluated Suite)} & \multicolumn{4}{l}{\textit{Human-verified multi-hop and historical reasoning suite reported in main experiments}} & \textbf{11,811} \\
\bottomrule
\end{tabular}%
}
\vspace{1mm}
\begin{flushleft}
\footnotesize{\textbf{\textit{Note:}} In the public release of \ourdataset, we additionally provide 1,647 single-hop direct concept recognition questions (139 for Text Concept QA, 1,032 for Visual Concept Recognition, and 476 for Concept VQA) as diagnostic baselines to monitor model fallback, bringing the total released benchmark to 13,458 QA pairs.}
\end{flushleft}
\end{table*}

\begin{figure}[h!]
    \centering
    \includegraphics[width=\linewidth]{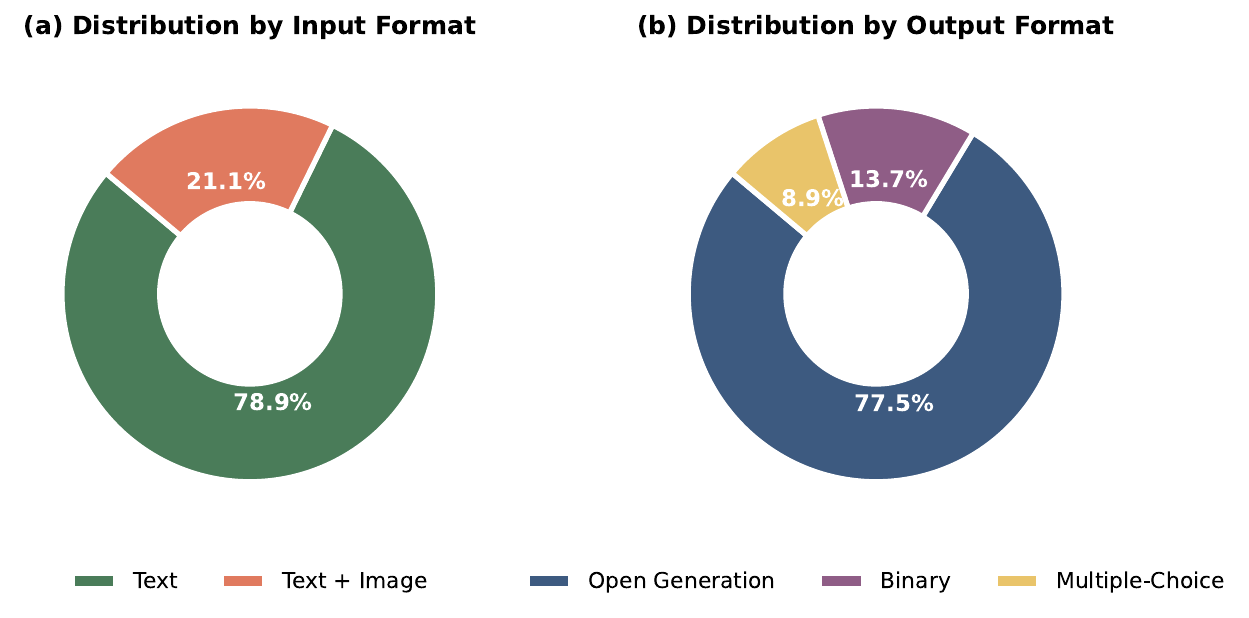}
    \caption{Distribution of the evaluated {\ourdataset} suite across (a) input formats and (b) output formats.}
    \label{fig:suppl_dataset_pies}
\end{figure}

This section provides more detailed of task specifications and benchmark statistics for {\ourdataset}.
\cref{tab:suppl_task_breakdown} summarizes the 10 evaluation tasks with their reasoning focus, input/output formats, and question counts;
\cref{fig:suppl_dataset_pies} shows the overall distribution across input and output modalities.

\subsection{Benchmark composition.}
The question count per task reflects the underlying personal archive structure.
Event-grounded tasks draw from hundreds of timestamped moments and images per Vaccount, yielding a large question pool, whereas concept- and persona-level tasks are bounded by the small social network (3--5 concepts per account) and longitudinal behavioral patterns.
The core evaluated benchmark comprises \textbf{11,811} human-verified question--answer pairs focusing on multi-hop relational reasoning, event localization, and history-wide aggregation.
We additionally release \textbf{1,647} single-hop direct concept questions across the three concept tasks, also with full human verification.
These are excluded from the primary evaluation because all evaluated retrieval approaches already near-saturate on single-hop concept lookup, limiting their discriminative value for method comparison; they are instead provided as a \emph{diagnostic set} for verifying that more complex personalization methods do not regress on basic entity recognition.
The total released collection comprises \textbf{13,458} QA pairs.

In addition to the human-verified evaluation QAs grounded in personal historical context (concepts, social networks, and multimodal event histories, grouped across 10 Vaccounts), we release a development set containing comparable-scale personal context data grouped across 40 additional Vaccounts. The development set is constructed through the same generation pipeline (\cref{subsec:dataset_constru}) and validated through multi-stage LLM verification (\cref{subsec:dataset_validation}), but without human annotation or task-specific QA pairs.
This DEV set provides a dedicated testbed for developing and tuning multimodal personalization methods without risk of overfitting to the evaluation set.

\subsection{Task Definitions}
\label{sec:suppl_task_definitions}

Section~3.2 introduces the 10 tasks in three categories and their required evidence scope.
Below we provide detail of each task, including the operational definition, reasoning requirements, and format for each of the 10 tasks.

\begin{itemize}[leftmargin=*]
    \item \textit{Text Concept QA}:
    Text-only queries over concept personas, relationship ties, and visual appearance.
    Multi-hop questions require traversing the social graph (\eg, identifying the account owner's maternal grandmother) and retrieving the resolved concept's attributes.

    \item \textit{Visual Concept Recognition}:
    Given a query image and a concept referred to by name or relation, the model must verify whether the concept appears in the image.
    Requires resolving relational chains and cross-modal identity matching against the concept's portrait profile.

    \item \textit{Concept VQA}:
    Fine-grained visual questions about a specific concept's attributes or actions within a provided image (\eg, what they are holding or wearing).
    Requires jointly resolving multi-hop social references and grounding visual evidence.

    \item \textit{Scene and Activity}:
    Holistic scene understanding and primary activity recognition for a dated event (\eg, event theme, environment, or key activity).
    Requires date-based or semantic event retrieval and multimodal context integration.

    \item \textit{Direct Person-Centric}:
    Fine-grained visual and behavioral details of a named concept during a specific historical event (\eg, attire, emotional expression, or physical interaction).
    Requires locating the target event and isolating the specified individual in its images.

    \item \textit{Relational Person-Centric}:
    Same as Direct Person-Centric, but the target individual is referred to via a relational description (\eg, ``the account owner's daughter'') rather than by name.
    Adds multi-hop social graph navigation before event-level visual reasoning.

    \item \textit{Fine-Grained Scene}:
    Compositional queries about localized interactions or subtle details within a specific moment (\eg, a background object or what another participant is doing during the primary action).
    Requires precise cross-modal frame localization.

    \item \textit{Preference and Persona}:
    Inferring stable preferences, recurring habits, or personal traits by aggregating behavioral evidence across the user's history.
    Questions present candidate options (\eg, identifying the most suitable gift), requiring a justified natural-language response grounded in historical events.

    \item \textit{Frequency and Counting}:
    Exhaustive quantitative aggregation over the complete timeline: the model must retrieve all unique events satisfying multi-condition criteria (\eg, counting how many times a specific person participated in outdoor activities) and output the precise count.

    \item \textit{Relational Temporal Reasoning}:
    Multi-step chronological reasoning: locate an anchor event via descriptive semantic clues, navigate backward or forward along the timeline to a relative event, and answer a question about that target event.
\end{itemize}

\subsection{Seeds Pool for {\ourdataset} Construction}
\label{suppl:yfcc_seeds}

To ground {\ourdataset}'s synthetic personal histories in realistic scenarios,
we derive textual seeds from the YFCC100M dataset~\cite{thomee2016yfcc100m}.
We filter YFCC100M to retain a subset of images released under Creative Commons (CC~BY) licenses, yielding over 169{,}000 images from over 400 Flickr accounts.
For each image, we generate a detailed, descriptive caption using
Gemini~2.5~Pro~\cite{google25gemini25}, producing rich depictions of subjects, activities, scenes, and visual details.

We construct two seed pools from these captions:
\textbf{Concept seeds} are captions centered on a single person or animal,
each describing their appearance, attire, and distinguishing attributes;
this pool contains 1{,}296 seeds.
\textbf{Event seeds} are formed by grouping the remaining captions by
YFCC account and date, then prompting the LLM to synthesize each group
into a coherent event narrative with a title, story, participating characters,
and per-image descriptions.
The resulting event-seed pool contains 8{,}681 events spanning 167{,}823
image-level captions.
These two pools supply the realistic subjects and event scenarios
that populate each Vaccount during benchmark construction (Section~3.3);
the original YFCC images are used solely as captioning inputs and not used in benchmark construction pipeline.
The descriptive captions from YFCC100M will not be released in benchmark, for privacy preserving.

\section{{\lifegraph} Details}
\label{sec:suppl_lifegraph}

This section provide more details of {\lifegraph}.

\subsection{Construction details}

Section~4.2 defines {\lifegraph}'s hybrid schema with six predefined entity types and open-ended relations.
The six entity types are specified as follows:
\begin{itemize}
    \item \emph{PersonAnimal} represents core personal concepts (people and pets);
    \item \emph{Event} represents a specific occurrence in the user's history, typically associated with a date and one or more images;
    \item \emph{Date} represents a specific date or timestamp;
    \item \emph{Location} represents a geographical place associated with an event or person;
    \item \emph{Activity} represents an action or occasion within an event;
    \item \emph{Object} covers any other descriptive facts (\eg, possessions, instruments, or foods) related to a concept or event.
\end{itemize}

To illustrate the $n$-ary hyper-relational encoding introduced in Section~4.2, consider the fact ``In the Christmas Party event, David and Rylen are decorating the Christmas tree using red ornaments.''
This is stored as a single triple with edge attributes: \texttt{Christmas Party event -- hasActivity\textless attendee: [David, Rylen], object: red ornaments\textgreater\ -- Decorate Christmas Tree},
preserving multi-argument structure without auxiliary nodes.

During the second construction stage (event history extraction), multimodal history records are processed in batches.
We employ a greedy chronological bin-packing strategy: sequential events are consolidated into batches whose aggregate size stays within the VLM's effective reasoning capacity rather than filling its full context window, balancing construction throughput with extraction quality. We use batch size 5 in the experiment.

\begin{figure}[h]
    \centering
    \includegraphics[width=0.8\linewidth]{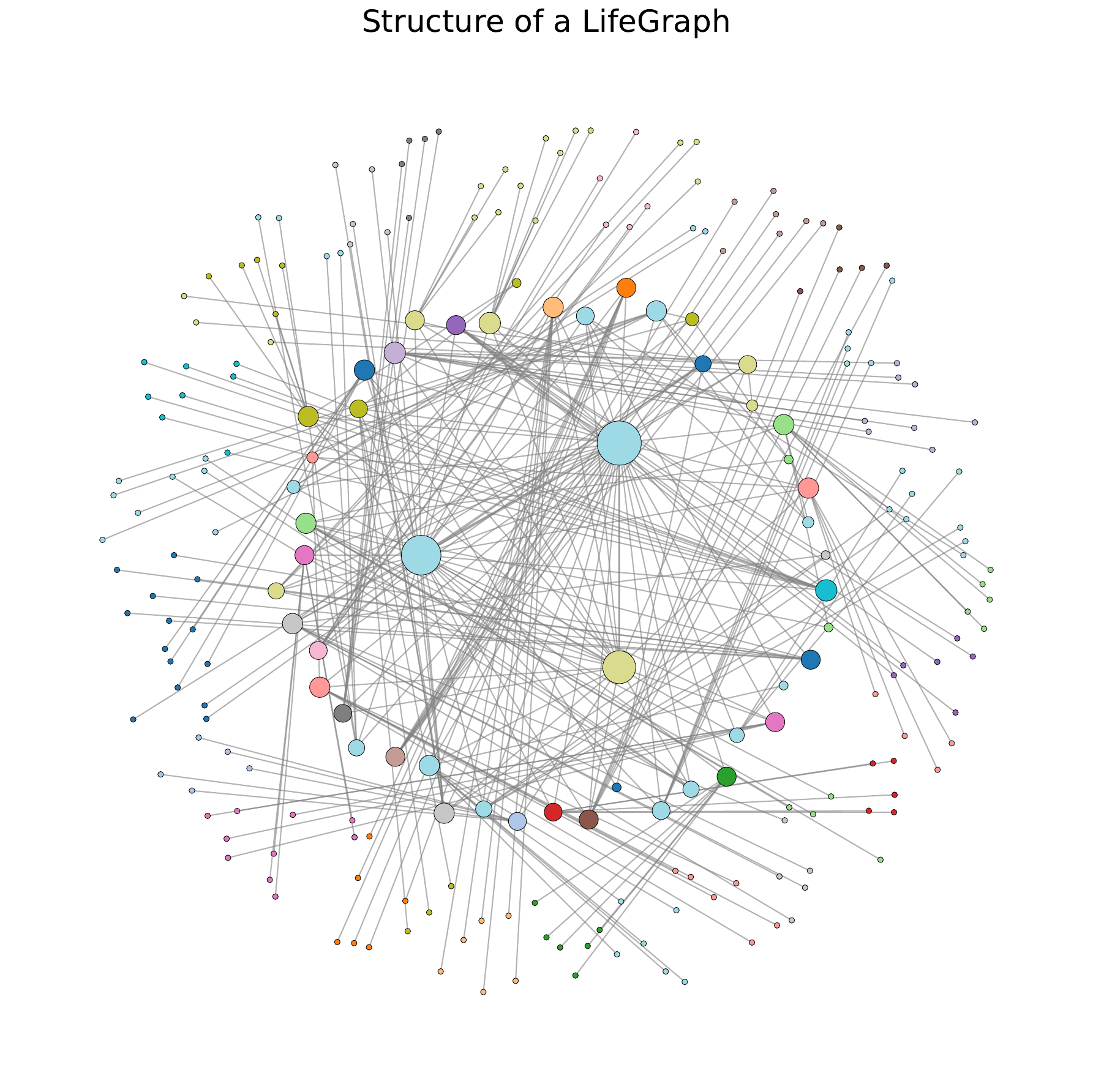}
    \caption{Graph structure of a constructed {\lifegraph}.}
    \label{fig:suppl_lifegraph_vis}
\end{figure}

\cref{fig:suppl_lifegraph_vis} visualizes a complete {\lifegraph} constructed from a {\ourdataset} Vaccount.

\subsection{Retrieval details}

\begin{figure}[h]
    \centering
    \includegraphics[width=0.95\linewidth]{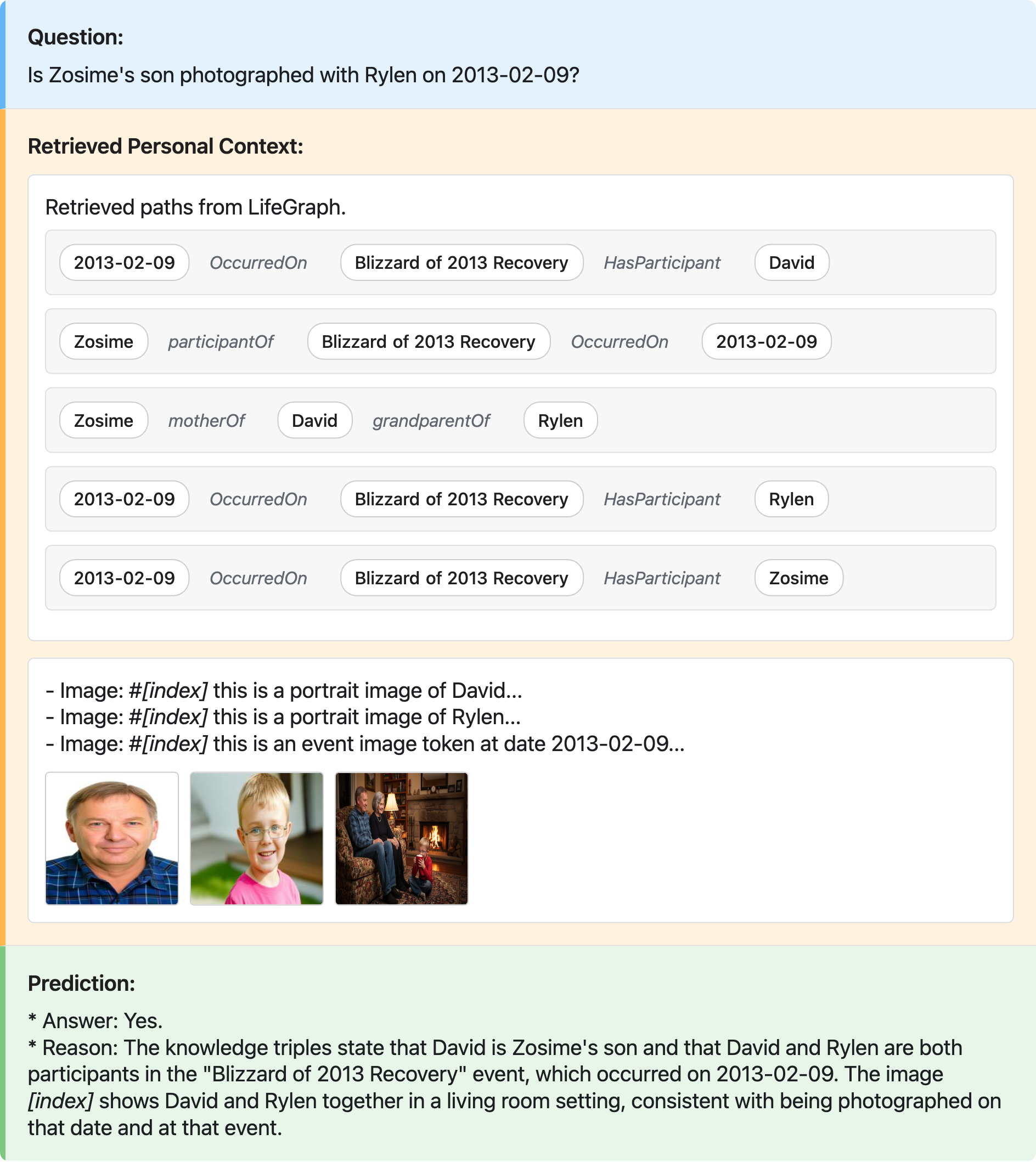}
    \caption{An example of the {\lifegraph} retrieval and reasoning process.}
    \label{fig:suppl_lifegraph_example}
\end{figure}

\begin{figure}[h]
    \centering
    \includegraphics[width=0.7\linewidth]{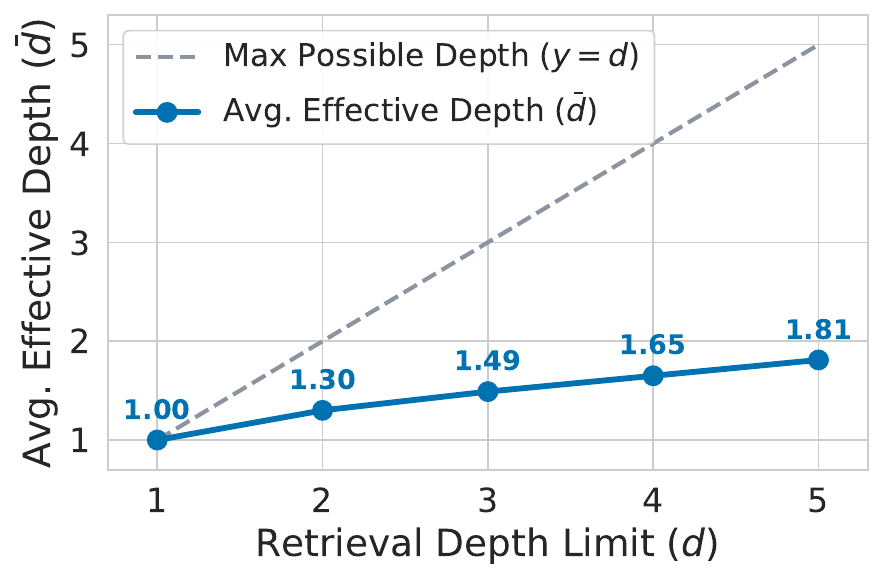}
    \caption{Average effective retrieve depth ($\bar{d}$) of {\lifegraph} under different retrieval depth limit ($d$).}
    \label{fig:avg_retrieve_depth}
\end{figure}

While {\lifegraph} stores directed edges to represent asymmetric relations, the retriever treats them as undirected for bidirectional traversal.
Retrieval initiates from multiple query-relevant top entities, allowing paths to be explored from different entry points.
In practice, the average effective depth remains below 2 even when the depth limit is raised to 5 (\cref{fig:avg_retrieve_depth}), indicating that most queries are resolved within a small neighborhood of the entry entities.
\cref{fig:suppl_lifegraph_example} provides a concrete walkthrough of the retrieval and reasoning process described in Algorithm~1, showing how traversed graph paths and fetched multimodal source references are combined to answer a personal query.

\section{Additional Experiment Results}

\subsection{LLM-as-Judge Agreement Details}

\begin{table}[h]
\centering
\caption{Per-method accuracy under GPT-5.4 vs.\ Gemini 3.6 Flash judges on the same 2,090 instances. 
Negative $\Delta$ indicates GPT-5.4 assigns lower accuracy (stricter grading).
The method ranking is identical under both judges.}
\label{tab:suppl_judge_by_method}
\resizebox{\linewidth}{!}{%
\begin{tabular}{lccccc}
\toprule
\textbf{Method} & $N$ & \textbf{GPT-5.4}($\%$) & \textbf{Gemini}($\%$) & $\Delta(\%)$ & \textbf{Agree(\%)} \\
\midrule
Closed-book               & 418 & 23.0 & 23.4 & $-$0.4 & 99.0 \\
RAP-Gemma$_{k{=}5}$       & 418 & 28.7 & 30.9 & $-$2.2 & 96.9 \\
RAG-Cap$_{k{=}5}$         & 418 & 34.4 & 40.7 & $-$6.2 & 93.8 \\
LifeGraph$_{k{=}5,d{=}2}$ & 418 & 41.1 & 45.2 & $-$4.1 & 93.5 \\
Oracle Flash               & 418 & 86.4 & 87.8 & $-$1.4 & 97.6 \\
\bottomrule
\end{tabular}%
}
\end{table}

\begin{table}[h]
\centering
\caption{Per-task agreement between Gemini 3.6 Flash and GPT-5.4 judges on 2,090 uniformly sampled open-ended instances.
Three concept identification tasks use exact-match scoring and are excluded.}
\label{tab:suppl_judge_agreement}
\small
\begin{tabular}{lccc}
\toprule
\textbf{Task} & $N$ & \textbf{Agree} & \textbf{Agree \%} \\
\midrule
Scene \& Activity          & 300 & 285 & 95.0 \\
Direct Person-Centric      & 300 & 287 & 95.7 \\
Relational Person-Centric  & 300 & 268 & 89.3 \\
Fine-Grained Scene         & 300 & 295 & 98.3 \\
Preference \& Persona      & 300 & 288 & 96.0 \\
Frequency \& Counting      & 295 & 295 & 100.0 \\
Relational Temporal        & 295 & 292 & 99.0 \\
\midrule
\textbf{Overall}            & \textbf{2,090} & \textbf{2,010} & \textbf{96.2} \\
\bottomrule
\end{tabular}
\end{table}

To validate the automatic evaluation, we uniformly sample 2,090 open-ended instances stratified across all 7 generation tasks, 10 Vaccounts, and 5 representative evaluation settings (Closed-book, RAG-Cap, RAP-Gemma, LifeGraph, and Oracle).
Each instance is independently scored as \textsc{Correct} or \textsc{Wrong} by both Gemini 3.6 Flash (the primary judge used throughout the paper) and GPT-5.4~\cite{gpt54}, using an identical evaluation prompt.
Three concept identification tasks use exact-match scoring and are excluded.
\cref{tab:suppl_judge_agreement} reports per-task agreement:
the two judges agree on 96.2\% of instances overall, with per-task rates ranging from 89.3\% (Relational Person-Centric) to 100\% (Frequency and Counting).
\cref{tab:suppl_judge_by_method} breaks down accuracy by method under each judge.
GPT-5.4 is consistently stricter, assigning lower accuracy across all methods, but the full method ranking is preserved.
Notably, agreement is at least 97.6\% for both Closed-book and Oracle Gemini-3.5-Flash, and the gap between LifeGraph and RAG-Cap is larger under GPT-5.4 than under Gemini, indicating that Gemini 3.6 Flash as judge does not exhibit self-preference bias toward Gemini-family responses.

\subsection{Per-task Breakdown of Vaccount}

\begin{figure*}[t]
    \centering
    \includegraphics[width=0.95\linewidth]{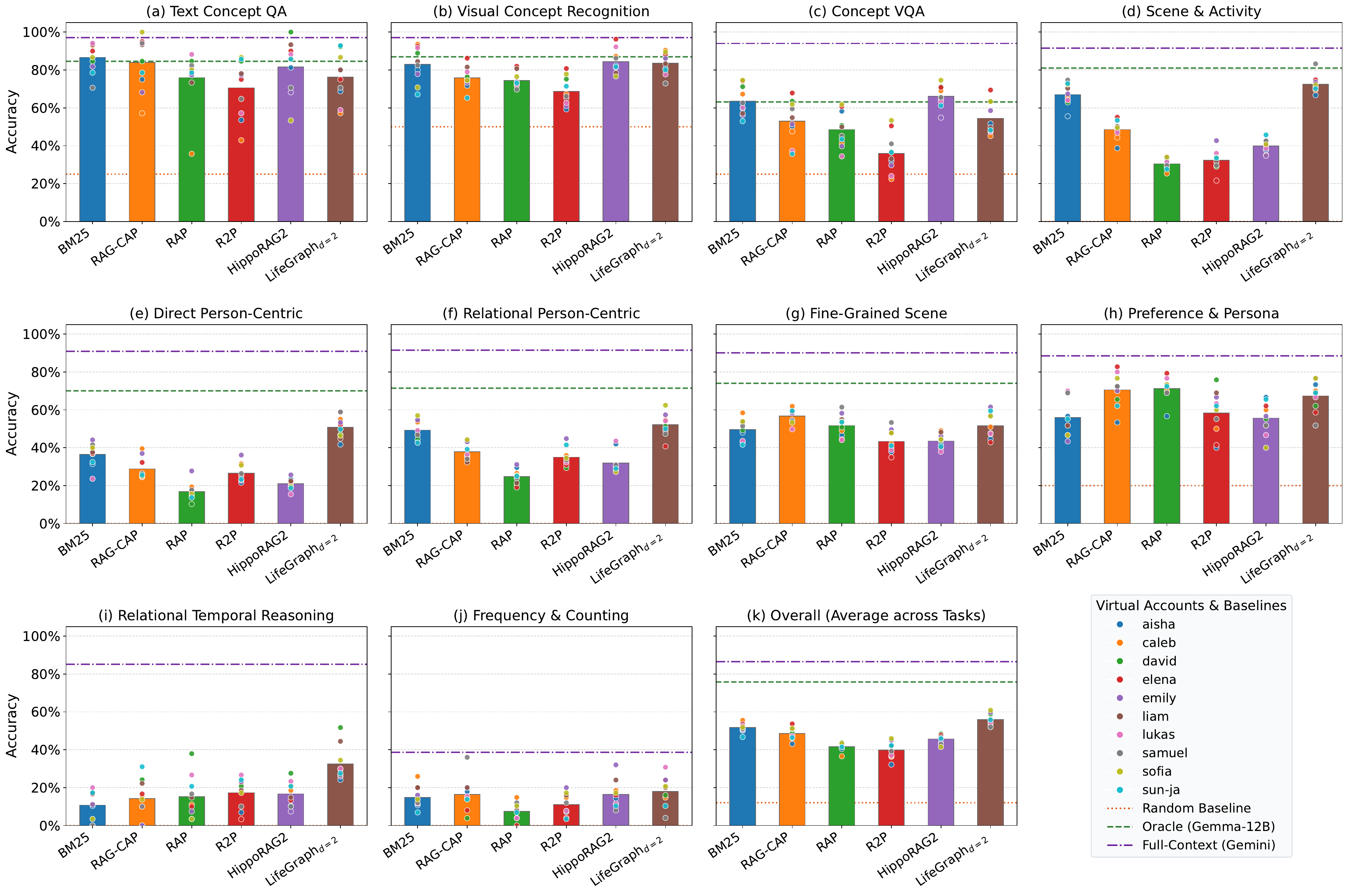}
    \caption{Per-Vaccount accuracy of retrieval methods at $k=5$ (LifeGraph: $d=2$), break down by per task. Bars show the mean across 10 Vaccounts, dots show individual Vaccounts.}
    \label{fig:suppl_vaccount_task_hist}
\end{figure*}

\begin{table*}[t]
\centering
\caption{Evaluation results across all Life-Bench tasks reported as $\text{mean} \pm \text{std}$ across 10 VAccounts ($N=10$). Retrieval-based methods are compared for context size $k \in \{3, 5\}$. Tasks are grouped into 3 categories. The best and second-best scores among each setting are in \textbf{bold} and \underline{underlined} respectively. ($^*$Oracle overall averages over seven concept and event tasks since aggregate does not have subset golden reference)}
\label{tab:suppl_main_results_std}
\resizebox{\textwidth}{!}{%
\begin{tabular}{l|ccc|cccc|ccc||c}
\toprule
\multirow{2}{*}{Method} & \multicolumn{3}{c|}{Concept} & \multicolumn{4}{c|}{Event} & \multicolumn{3}{c||}{Aggregated} & \multirow{2}{*}{Overall} \\
\cmidrule(lr){2-4} \cmidrule(lr){5-8} \cmidrule(lr){9-11}
 & \begin{tabular}[c]{@{}c@{}}Text Concept\\QA\end{tabular} & \begin{tabular}[c]{@{}c@{}}Visual Concept\\Recognition\end{tabular} & \begin{tabular}[c]{@{}c@{}}Concept\\VQA\end{tabular} & \begin{tabular}[c]{@{}c@{}}Scene \&\\Activity\end{tabular} & \begin{tabular}[c]{@{}c@{}}Direct\\Person-Centric\end{tabular} & \begin{tabular}[c]{@{}c@{}}Relational\\Person-Centric\end{tabular} & \begin{tabular}[c]{@{}c@{}}Fine-Grained\\Scene\end{tabular} & \begin{tabular}[c]{@{}c@{}}Preference\\\& Persona\end{tabular} & \begin{tabular}[c]{@{}c@{}}Relational\\Temporal\end{tabular} & \begin{tabular}[c]{@{}c@{}}Frequency\\\& Counting\end{tabular} & \\
\midrule
Full-context & 0.9707\std{0.039} & 0.9712\std{0.012} & 0.9403\std{0.052} & 0.9145\std{0.012} & 0.9091\std{0.027} & 0.9146\std{0.043} & 0.9012\std{0.021} & 0.8853\std{0.058} & 0.8517\std{0.082} & 0.3871\std{0.144} & 0.8646\std{0.017} \\
Oracle & 0.9786\std{0.048} & 0.9815\std{0.015} & 0.9525\std{0.038} & 0.9318\std{0.020} & 0.9286\std{0.032} & 0.9271\std{0.035} & 0.9100\std{0.026} & -- & -- & -- & 0.9443$^*$\std{0.012} \\
Oracle-Gemma & 0.8457\std{0.099} & 0.8691\std{0.063} & 0.6315\std{0.109} & 0.8093\std{0.023} & 0.7005\std{0.035} & 0.7138\std{0.082} & 0.7401\std{0.045} & -- & -- & -- & 0.7586$^*$\std{0.039} \\
\midrule
$\text{BM25}_{k=3}$  & \underline{0.8138}\std{0.112} & 0.8139\std{0.073} & \underline{0.5850}\std{0.066} & 0.6160\std{0.063} & 0.3416\std{0.062} & 0.4416\std{0.069} & 0.4738\std{0.044} & 0.4784\std{0.105} & 0.1179\std{0.075} & 0.1345\std{0.073} & 0.4816\std{0.025} \\
$\text{RAG-Cap}_{k=3}$  & \textbf{0.8205}\std{0.148} & 0.7366\std{0.054} & 0.5252\std{0.107} & 0.4369\std{0.039} & 0.2515\std{0.032} & 0.3376\std{0.036} & \textbf{0.5253}\std{0.040} & 0.6374\std{0.067} & 0.1249\std{0.086} & 0.1397\std{0.067} & 0.4535\std{0.034} \\
$\text{RAP}_{k=3}$ & 0.4513\std{0.098} & 0.3976\std{0.052} & 0.4478\std{0.078} & 0.3316\std{0.028} & 0.2227\std{0.035} & 0.3350\std{0.046} & 0.3074\std{0.039} & 0.5525\std{0.082} & 0.1603\std{0.068} & 0.0943\std{0.035} & 0.3301\std{0.028} \\
$\text{RAP-Gemma}_{k=3}$ & 0.7526\std{0.189} & 0.7399\std{0.056} & 0.4969\std{0.110} & 0.2748\std{0.029} & 0.1248\std{0.029} & 0.2088\std{0.042} & 0.4425\std{0.051} & \textbf{0.6714}\std{0.084} & 0.1170\std{0.063} & 0.0939\std{0.038} & 0.3923\std{0.036} \\
$\text{R2P}_{k=3}$  & 0.6214\std{0.174} & 0.5442\std{0.045} & 0.3764\std{0.075} & 0.2630\std{0.025} & 0.2006\std{0.043} & 0.2934\std{0.059} & 0.3865\std{0.045} & 0.4682\std{0.131} & 0.1538\std{0.057} & 0.0452\std{0.024} & 0.3353\std{0.029} \\
$\text{R2P-Gemma}_{k=3}$  & 0.6406\std{0.141} & 0.6845\std{0.060} & 0.3860\std{0.117} & 0.3061\std{0.044} & 0.2658\std{0.054} & 0.3430\std{0.062} & 0.3943\std{0.059} & 0.5616\std{0.121} & 0.1717\std{0.078} & 0.0679\std{0.039} & 0.3822\std{0.036} \\
$\text{HippoRAG2}_{k=3}$  & 0.8129\std{0.083} & 0.8056\std{0.074} & \textbf{0.6182}\std{0.078} & 0.3363\std{0.020} & 0.1762\std{0.019} & 0.2271\std{0.067} & 0.3368\std{0.040} & 0.4848\std{0.085} & 0.1428\std{0.090} & 0.1255\std{0.056} & 0.4066\std{0.018} \\
\hdashline
$\text{LifeGraph}_{d=2,k=3}$ & 0.7094\std{0.154} & \textbf{0.8274}\std{0.058} & 0.5521\std{0.086} & \textbf{0.7139}\std{0.056} & \textbf{0.4557}\std{0.065} & \textbf{0.5252}\std{0.077} & 0.5073\std{0.070} & \underline{0.6574}\std{0.080} & \textbf{0.2933}\std{0.076} & \textbf{0.1849}\std{0.052} & \textbf{0.5427}\std{0.032} \\
$\text{LifeGraph}_{d=3,k=3}$ & 0.7767\std{0.099} & \underline{0.8258}\std{0.056} & 0.5546\std{0.107} & \underline{0.7095}\std{0.066} & \underline{0.4471}\std{0.064} & \underline{0.5083}\std{0.063} & \underline{0.5100}\std{0.058} & 0.6469\std{0.092} & \underline{0.2821}\std{0.088} & \underline{0.1522}\std{0.071} & \underline{0.5413}\std{0.029} \\
\midrule
\midrule
$\text{BM25}_{k=5}$  & \textbf{0.8663}\std{0.078} & 0.8292\std{0.091} & \underline{0.6357}\std{0.078} & 0.6701\std{0.058} & 0.3647\std{0.060} & 0.4928\std{0.054} & 0.4957\std{0.055} & 0.5599\std{0.100} & 0.1072\std{0.074} & 0.1484\std{0.057} & 0.5170\std{0.024} \\
$\text{RAG-Cap}_{k=5}$  & \underline{0.8401}\std{0.140} & 0.7579\std{0.057} & 0.5296\std{0.108} & 0.4849\std{0.046} & 0.2884\std{0.054} & 0.3789\std{0.043} & \textbf{0.5675}\std{0.037} & \underline{0.7052}\std{0.087} & 0.1427\std{0.100} & 0.1635\std{0.086} & 0.4859\std{0.032} \\
$\text{RAP-Gemma}_{k=5}$  & 0.7596\std{0.147} & 0.7449\std{0.040} & 0.4853\std{0.094} & 0.3046\std{0.028} & 0.1679\std{0.046} & 0.2470\std{0.038} & 0.5168\std{0.058} & \textbf{0.7121}\std{0.061} & 0.1522\std{0.109} & 0.0745\std{0.046} & 0.4165\std{0.021} \\
$\text{R2P-Gemma}_{k=5}$  & 0.7056\std{0.152} & 0.6865\std{0.073} & 0.3593\std{0.102} & 0.3233\std{0.054} & 0.2658\std{0.047} & 0.3497\std{0.048} & 0.4331\std{0.060} & 0.5834\std{0.117} & 0.1728\std{0.079} & 0.1100\std{0.058} & 0.3989\std{0.043} \\
$\text{HippoRAG2}_{k=5}$  & 0.8163\std{0.138} & \textbf{0.8437}\std{0.061} & \textbf{0.6603}\std{0.057} & 0.3986\std{0.034} & 0.2092\std{0.028} & 0.3193\std{0.059} & 0.4355\std{0.036} & 0.5562\std{0.084} & 0.1667\std{0.065} & 0.1635\std{0.072} & 0.4569\std{0.023} \\
\hdashline
$\text{LifeGraph}_{d=2,k=5}$ & 0.7624\std{0.126} & \underline{0.8358}\std{0.057} & 0.5444\std{0.085} & \textbf{0.7240}\std{0.044} & \textbf{0.5078}\std{0.059} & \underline{0.5211}\std{0.059} & 0.5154\std{0.066} & 0.6737\std{0.080} & \underline{0.3246}\std{0.089} & \underline{0.1794}\std{0.076} & \underline{0.5589}\std{0.031} \\
$\text{LifeGraph}_{d=3,k=5}$ & 0.7919\std{0.133} & 0.8304\std{0.065} & 0.5384\std{0.098} & \underline{0.7227}\std{0.050} & \underline{0.4970}\std{0.058} & \textbf{0.5364}\std{0.062} & \underline{0.5252}\std{0.052} & 0.6513\std{0.095} & \textbf{0.3268}\std{0.115} & \textbf{0.1987}\std{0.072} & \textbf{0.5619}\std{0.032} \\
\bottomrule
\end{tabular}%
}
\end{table*}

\cref{fig:vaccount_category_hist} in the main paper reports per-Vaccount accuracy grouped by task category.
Figure~\ref{fig:suppl_vaccount_task_hist} provides a finer-grained view, breaking down accuracy by individual task for each retrieval method. 
To further quantify variance across Vaccounts, \cref{tab:suppl_main_results_std} details the complete evaluation matrix with cross-Vaccount standard deviations ($\text{mean} \pm \text{std}$, $N=10$). 
Note that while \cref{tab:main_results_merged} in the main paper reports pooled accuracy per task, \cref{tab:suppl_main_results_std} reports the cross-Vaccount macro $\text{mean} \pm \text{std}$ ($N=10$) to characterize user-level dispersion; while numerical means differ slightly due to this grouping, all relative rankings and conclusions remain identical.
The per-Vaccount patterns observed at the category level carry over to individual tasks: method rankings remain largely consistent across Vaccounts, and the cross-Vaccount variance within each method is substantially smaller than the gaps between methods.

\subsection{Per-task Recall}
In \cref{tab:recall}, we reported recall across methods at the category level. \Cref{tab:per_task_recall} provides the detailed per-task breakdown. Note that Aggregated Reasoning tasks are omitted from recall evaluation because they require cross-history synthesis across the entire timeline rather than retrieving specific localized gold events.

\begin{table*}[t]
\centering
\caption{Per-task event-level Recall@$k$ on Life-Bench. Retrieving any image from a gold event counts as a hit.
The best and second-best scores per column are in \textbf{bold} and \underline{underlined}, respectively.}
\label{tab:per_task_recall}
\resizebox{\textwidth}{!}{%
\begin{tabular}{l cc cc cc cc cc cc cc}
\toprule
& \multicolumn{6}{c}{\textbf{Concept}} & \multicolumn{8}{c}{\textbf{Event}} \\
\cmidrule(lr){2-7} \cmidrule(lr){8-15}
& \multicolumn{2}{c}{Text Concept QA} & \multicolumn{2}{c}{Visual Concept} & \multicolumn{2}{c}{Concept VQA}
& \multicolumn{2}{c}{Scene \& Activity} & \multicolumn{2}{c}{Direct Person} & \multicolumn{2}{c}{Rel.\ Person} & \multicolumn{2}{c}{Fine-Grained Scene} \\
\cmidrule(lr){2-3} \cmidrule(lr){4-5} \cmidrule(lr){6-7}
\cmidrule(lr){8-9} \cmidrule(lr){10-11} \cmidrule(lr){12-13} \cmidrule(lr){14-15}
Method & R@3 & R@5 & R@3 & R@5 & R@3 & R@5 & R@3 & R@5 & R@3 & R@5 & R@3 & R@5 & R@3 & R@5 \\
\midrule
BM25~\cite{robertson2009bm25}
  & \textbf{0.5456} & \textbf{0.6246}
  & 0.5157 & 0.5733
  & 0.3265 & 0.4010
  & \textbf{0.6021} & \textbf{0.6541}
  & 0.4989 & 0.5420
  & 0.4499 & 0.5329
  & 0.3629 & 0.4192 \\
RAG-Cap~\cite{lewis2020rag}
  & \underline{0.5404} & \underline{0.6193}
  & 0.4638 & 0.5048
  & 0.2879 & 0.3183
  & 0.3460 & 0.4164
  & 0.3703 & 0.4687
  & 0.2608 & 0.3486
  & 0.4890 & 0.5984 \\
R2P-Gemma~\cite{das2025r2p}
  & 0.3474 & 0.3781
  & 0.4348 & 0.4429
  & \underline{0.4388} & \underline{0.4505}
  & 0.1960 & 0.2358
  & 0.3841 & 0.4251
  & 0.3484 & 0.3889
  & \underline{0.6007} & \underline{0.6698} \\
RAP-Gemma~\cite{hao2025rap}
  & 0.3711 & 0.3982
  & 0.4582 & 0.5007
  & \textbf{0.4768} & \textbf{0.4969}
  & 0.2295 & 0.2714
  & 0.3807 & 0.4189
  & 0.3293 & 0.3741
  & \textbf{0.6863} & \textbf{0.7644} \\
HippoRAG2~\cite{gutierrez2025hipporag2}
  & 0.5333 & 0.5974
  & 0.5314 & 0.5858
  & 0.2748 & 0.3554
  & 0.2346 & 0.3147
  & 0.2725 & 0.3584
  & 0.3505 & 0.4565
  & 0.3591 & 0.4966 \\
\midrule
LifeGraph$_{d{=}2}$
  & 0.4123 & 0.4263
  & \underline{0.5906} & \textbf{0.6346}
  & 0.4213 & 0.4418
  & \underline{0.5779} & \underline{0.6432}
  & \textbf{0.5767} & \textbf{0.6307}
  & \underline{0.5072} & \underline{0.5552}
  & 0.4616 & 0.5793 \\
LifeGraph$_{d{=}3}$
  & 0.4237 & 0.4544
  & \textbf{0.6028} & \underline{0.5975}
  & 0.4267 & 0.4455
  & 0.5767 & 0.6370
  & \underline{0.5576} & \underline{0.6172}
  & \textbf{0.5073} & \textbf{0.5567}
  & 0.4667 & 0.5741 \\
\bottomrule
\end{tabular}%
}
\end{table*}

\subsection{Computational Cost and Efficiency}
\label{subsec:suppl_cost}

The retrieval methods evaluated on Life-Bench span a wide range of query-time computational cost. 
BM25~\cite{robertson2009bm25}, RAG-Cap~\cite{lewis2020rag}, and RAP~\cite{hao2025rap} require no VLM calls at retrieval time, relying entirely on pre-built lexical, embedding, or concept-memory indices to select context. 
R2P~\cite{das2025r2p} and HippoRAG2~\cite{gutierrez2025hipporag2} each invoke the VLM a fixed number of times per query for query analysis and candidate scoring, independent of traversal depth. 
{\lifegraph} incurs three VLM calls per traversal depth (Prune, FetchReference, and Reasoning) plus one initial call for topic entity extraction; its retrieval cost therefore grows linearly with depth, though the average effective depth remains below 2 in practice (\cref{fig:avg_retrieve_depth}). 
At the other extreme, the full-context diagnostic baseline bypasses retrieval entirely but processes over 200 images per query, consuming 30–50$\times$ more input tokens and cannot scale to real personal archives spanning thousands of photographs.
These cost profiles reveal a clear trade-off between retrieval cost and reasoning and retrieval accuracy. 
We believe that developing approaches that maintain the accuracy benefits of structured traversal at substantially lower inference cost is an important direction for future work, and we hope {\ourdataset} can serve as a testbed for measuring progress along both axes.

\section{Scope and Limitation}
\label{sec:suppl_dis}

{\ourdataset} evaluates personalization over complete but temporally bounded personal histories. Each Vaccount spans a coherent life period with consistent concept identities; the benchmark does not explicitly model long-horizon identity evolution such as changes in physical appearance due to aging or shifts in personal interests over decades, nor does it provide a continual-learning protocol for measuring how systems adapt to incrementally arriving records. 
Both are valuable research directions but fall outside the scope of the current work.
Relatedly, each Vaccount contains hundreds of images and dozens of events to support controlled evaluation across 10 tasks with full human verification; real personal archives typically accumulate far more photographs, even more than tens of thousands, over many years.
This difference reflects a practical tradeoff between benchmark scale, construction cost, and verification rigor, and we leave extending to larger-scale personal histories as future work.

Life-Bench group the multimodal personal content on 10 Vaccounts, a count shaped by the per-account personal content depth and diversity required to meaningfully evaluate cross-modal reasoning and history-wide aggregation. Each Vaccount maintains a realistically scaled archive: averaging 289 images across 38 events with a multi-concept social network—and the resulting 11,800+ verified QA pairs span 10 tasks covering diverse temporal, relational, and visual queries. 
Full-coverage human verification of every image and QA pair (\cref{subsec:dataset_validation}) further bounds the feasible account count at this depth. The 10 Vaccounts span varied social structures, event compositions, and personal contexts; per-Vaccount analysis (\cref{fig:vaccount_category_hist}) demonstrate that method rankings hold consistently across accounts, supporting the robustness of the benchmark's comparative conclusions at this scale. An additional 40 development Vaccounts are released to support method development at broader scale.

{\lifegraph} demonstrates that organizing personal knowledge as a structured, source-linked graph improves retrieval for event-level and aggregated tasks, yet it represents a directional exploration of this design space rather than a production-ready solution. 
As analyzed in \cref{subsec:suppl_cost}, its multi-call retrieval process incurs higher inference cost than flat retrieval baselines; 
developing methods that preserve the accuracy benefits of structured traversal at lower cost remains an important open problem. 
Additionally, as graph construction relies on general-purpose VLMs, advances in models with stronger relational and temporal extraction and understanding capabilities would directly benefit this approach. We hope that {\lifegraph}'s results encourage further exploration in this direction.

\section{Prompt Templates}
\label{sec:suppl_prompt}

\begin{figure*}[t]
  \centering
  \includegraphics[width=0.8\textwidth]{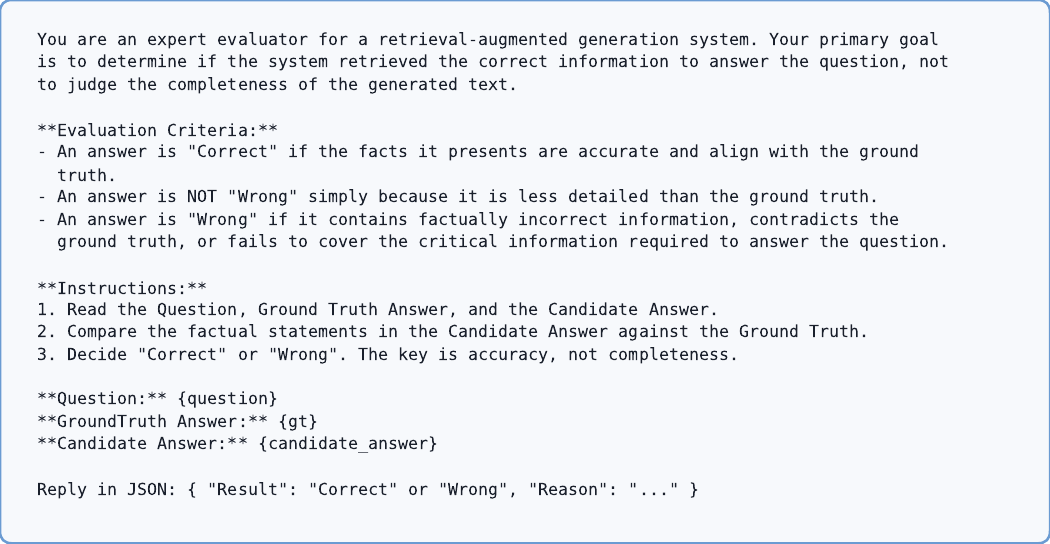}
  \caption{Evaluation Prompt for Auto-Rater (LLM-as-a-Judge).}
  \label{fig:prompt_autorater}
\end{figure*}
\begin{figure*}[t]
  \centering
  \includegraphics[width=0.8\textwidth]{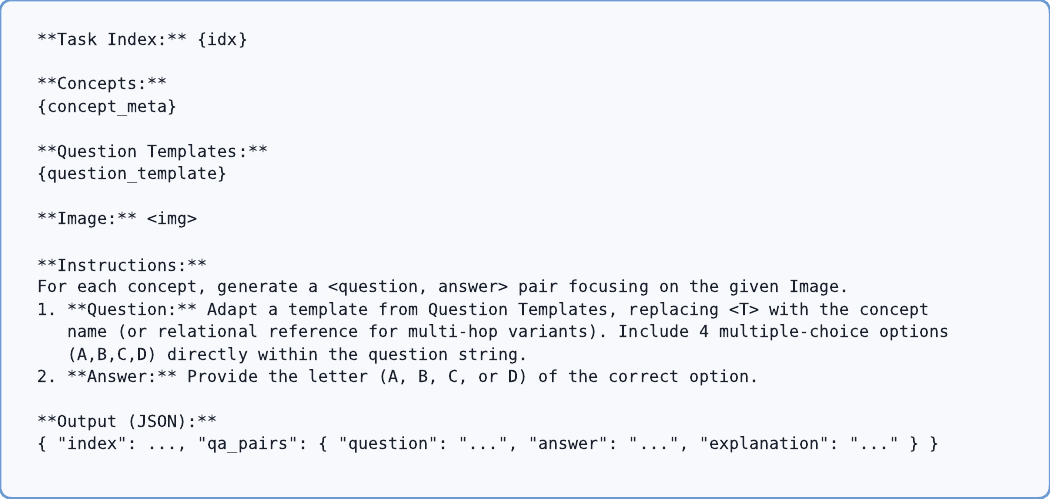}
  \caption{QA Generation Prompt Skeleton for Concept Identification Tasks.}
  \label{fig:prompt_skeleton_a}
\end{figure*}
\begin{figure*}[t]
  \centering
  \includegraphics[width=0.8\textwidth]{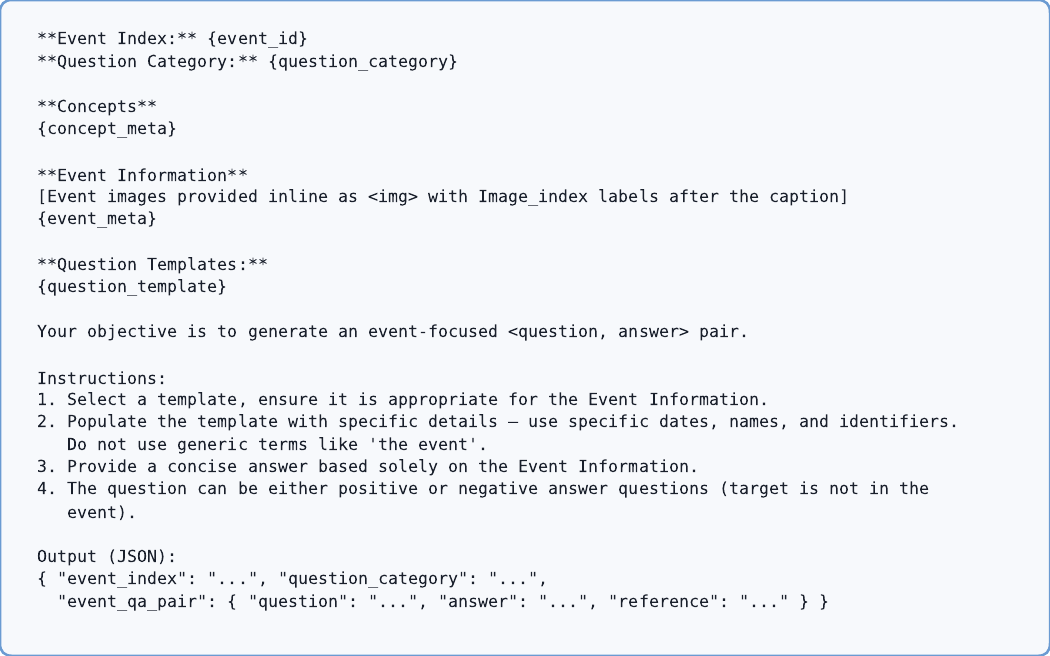}
  \caption{QA Generation Prompt Skeleton for Event Understanding Tasks.}
  \label{fig:prompt_skeleton_b}
\end{figure*}
\begin{figure*}[t]
  \centering
  \includegraphics[width=0.8\textwidth]{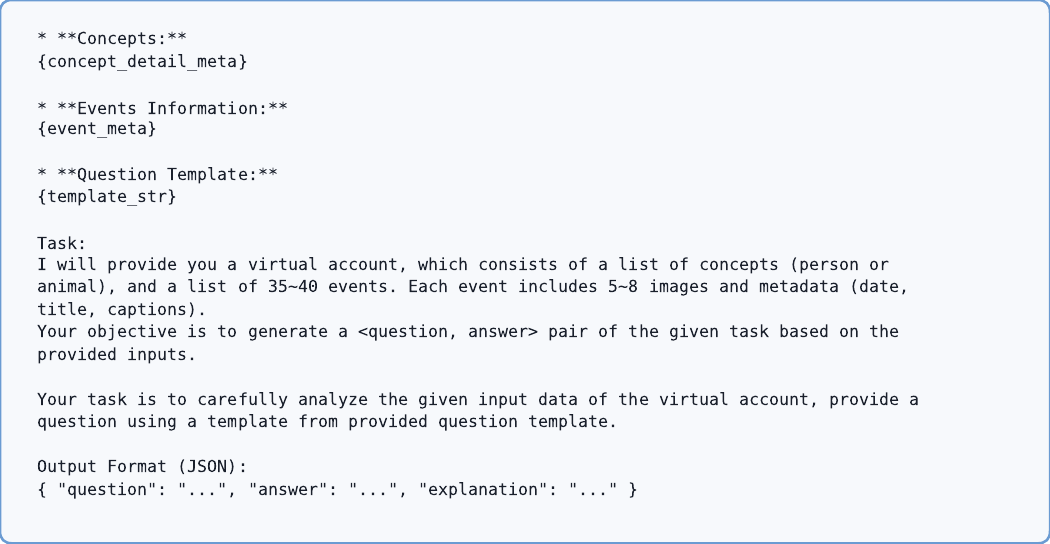}
  \caption{QA Generation Prompt Skeleton for Aggregated Reasoning Tasks.}
  \label{fig:prompt_skeleton_c}
\end{figure*}
\begin{figure*}[t]
  \centering
  \includegraphics[width=0.8\textwidth]{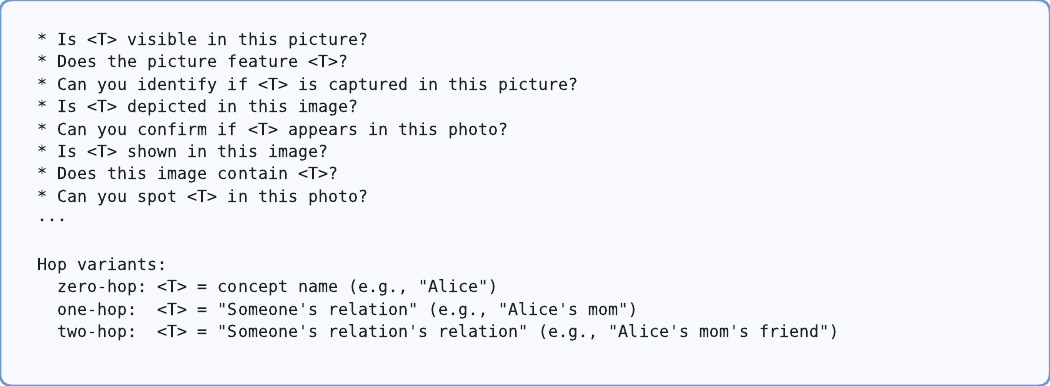}
  \caption{Question Template for Visual Concept Recognition.}
  \label{fig:prompt_tmpl_visual_recognition}
\end{figure*}
\begin{figure*}[t]
  \centering
  \includegraphics[width=0.8\textwidth]{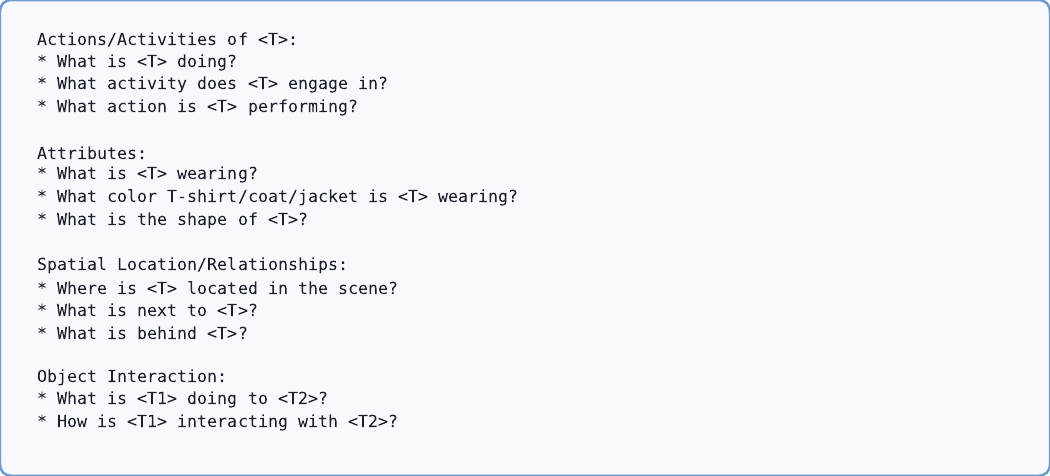}
  \caption{Question Template for Concept VQA.}
  \label{fig:prompt_tmpl_concept_vqa}
\end{figure*}
\begin{figure*}[t]
  \centering
  \includegraphics[width=0.8\textwidth]{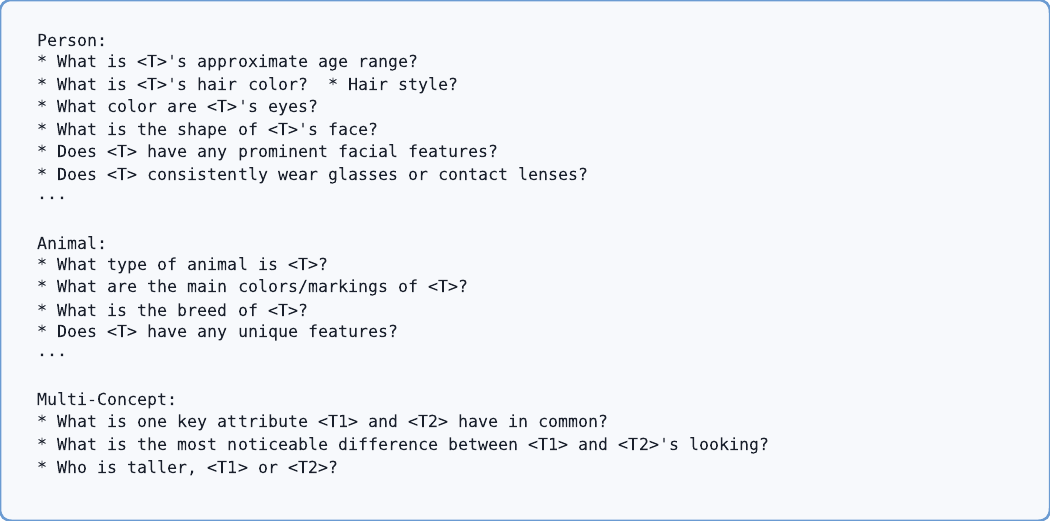}
  \caption{Question Template for Text Concept QA.}
  \label{fig:prompt_tmpl_text_concept_qa}
\end{figure*}
\begin{figure*}[t]
  \centering
  \includegraphics[width=0.8\textwidth]{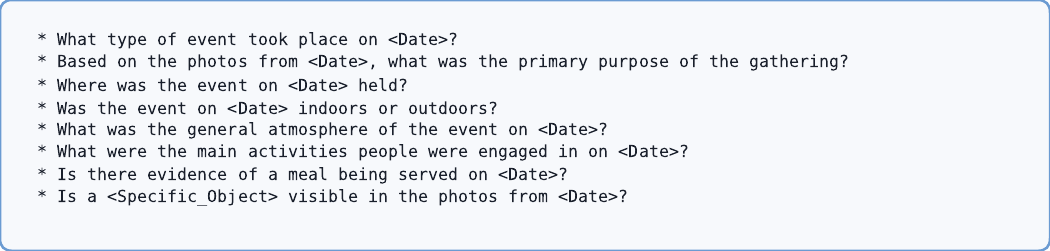}
  \caption{Question Template for Scene and Activity.}
  \label{fig:prompt_tmpl_scene_activity}
\end{figure*}
\begin{figure*}[t]
  \centering
  \includegraphics[width=0.8\textwidth]{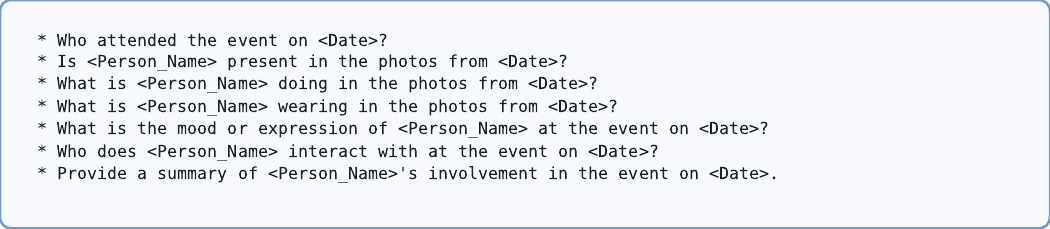}
  \caption{Question Template for Direct Person-Centric.}
  \label{fig:prompt_tmpl_direct_person}
\end{figure*}
\begin{figure*}[t]
  \centering
  \includegraphics[width=0.8\textwidth]{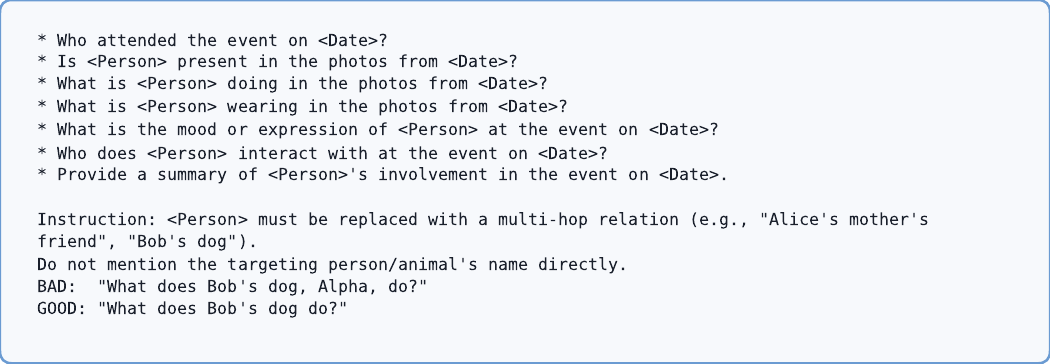}
  \caption{Question Template for Relational Person-Centric.}
  \label{fig:prompt_tmpl_relational_person}
\end{figure*}
\begin{figure*}[t]
  \centering
  \includegraphics[width=0.8\textwidth]{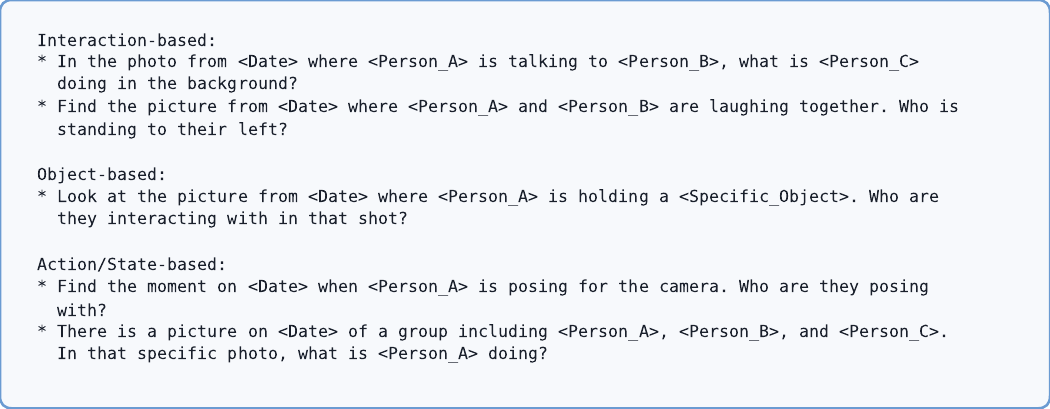}
  \caption{Question Template for Fine-Grained Scene.}
  \label{fig:prompt_tmpl_finegrained_scene}
\end{figure*}
\begin{figure*}[t]
  \centering
  \includegraphics[width=0.8\textwidth]{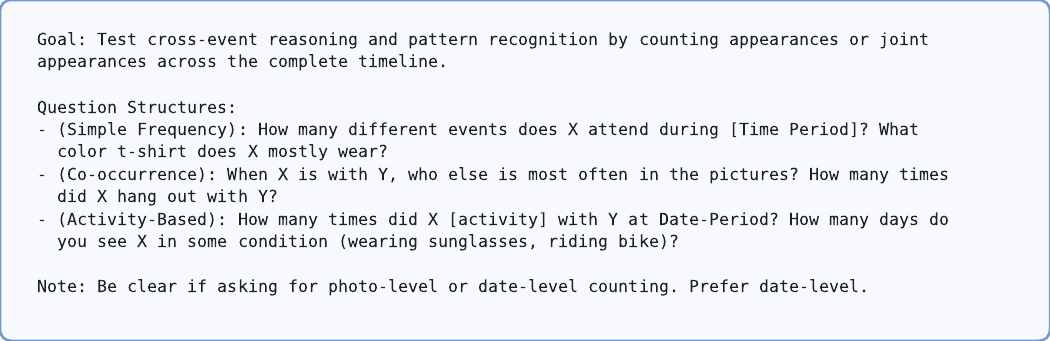}
  \caption{Question Template for Frequency and Counting.}
  \label{fig:prompt_tmpl_frequency_counting}
\end{figure*}
\begin{figure*}[t]
  \centering
  \includegraphics[width=0.8\textwidth]{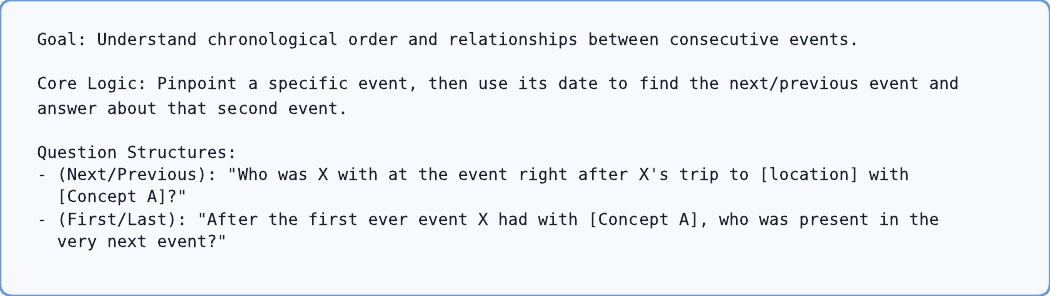}
  \caption{Question Template for Relational Temporal Reasoning.}
  \label{fig:prompt_tmpl_relational_temporal}
\end{figure*}
\begin{figure*}[t]
  \centering
  \includegraphics[width=0.8\textwidth]{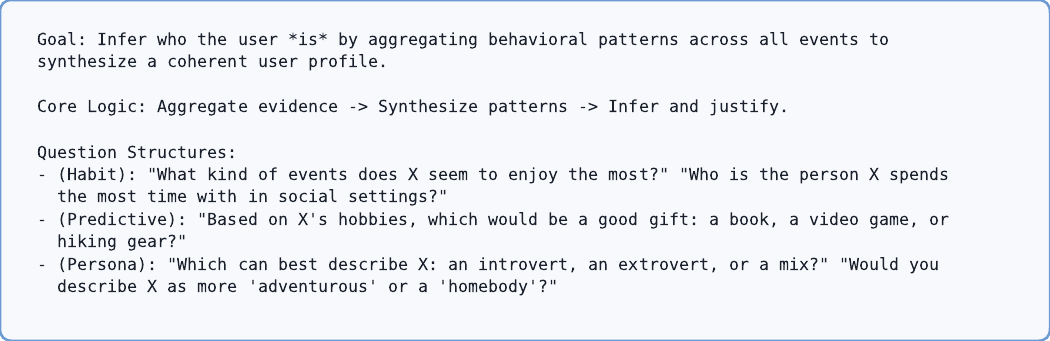}
  \caption{Question Template for Preference and Persona.}
  \label{fig:prompt_tmpl_preference_persona}
\end{figure*}
\begin{figure*}[t]
  \centering
  \includegraphics[width=0.8\textwidth]{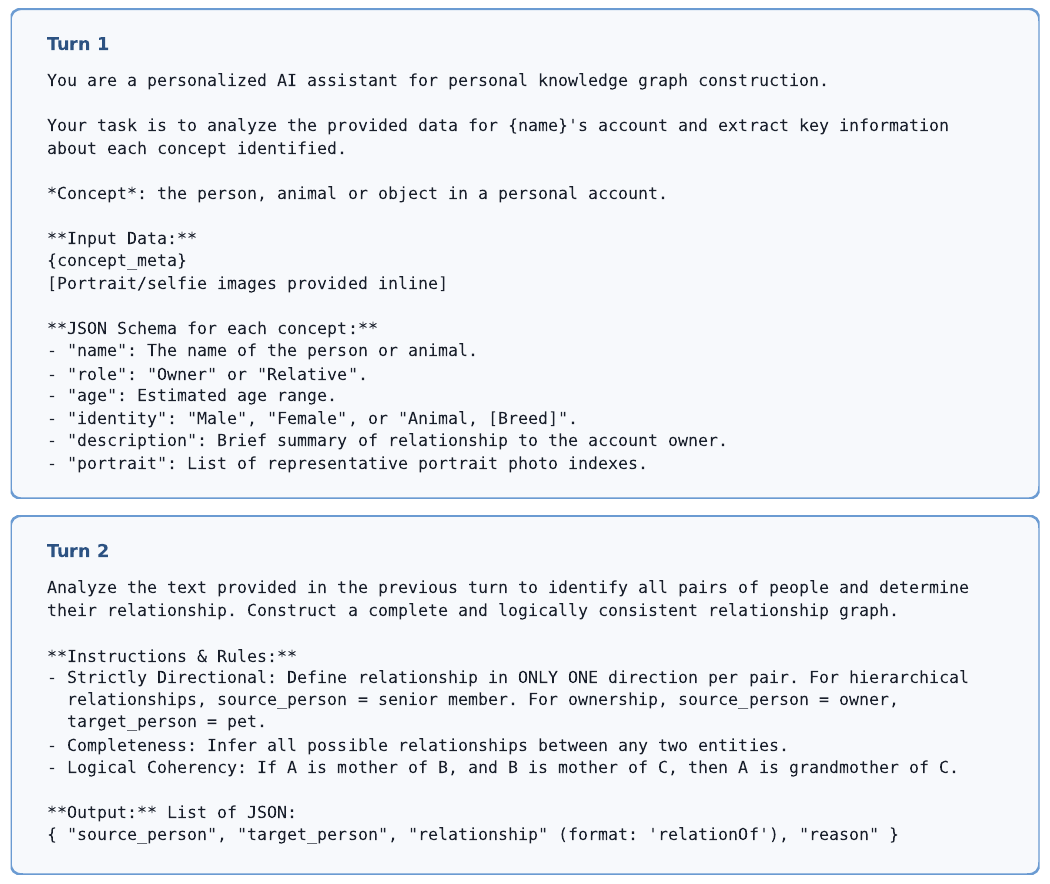}
  \caption{Two-Turn Prompt for LifeGraph Stage 1 (Concept Subgraph Construction).}
  \label{fig:prompt_lifegraph_stage1}
\end{figure*}
\begin{figure*}[t]
  \centering
  \includegraphics[width=0.8\textwidth]{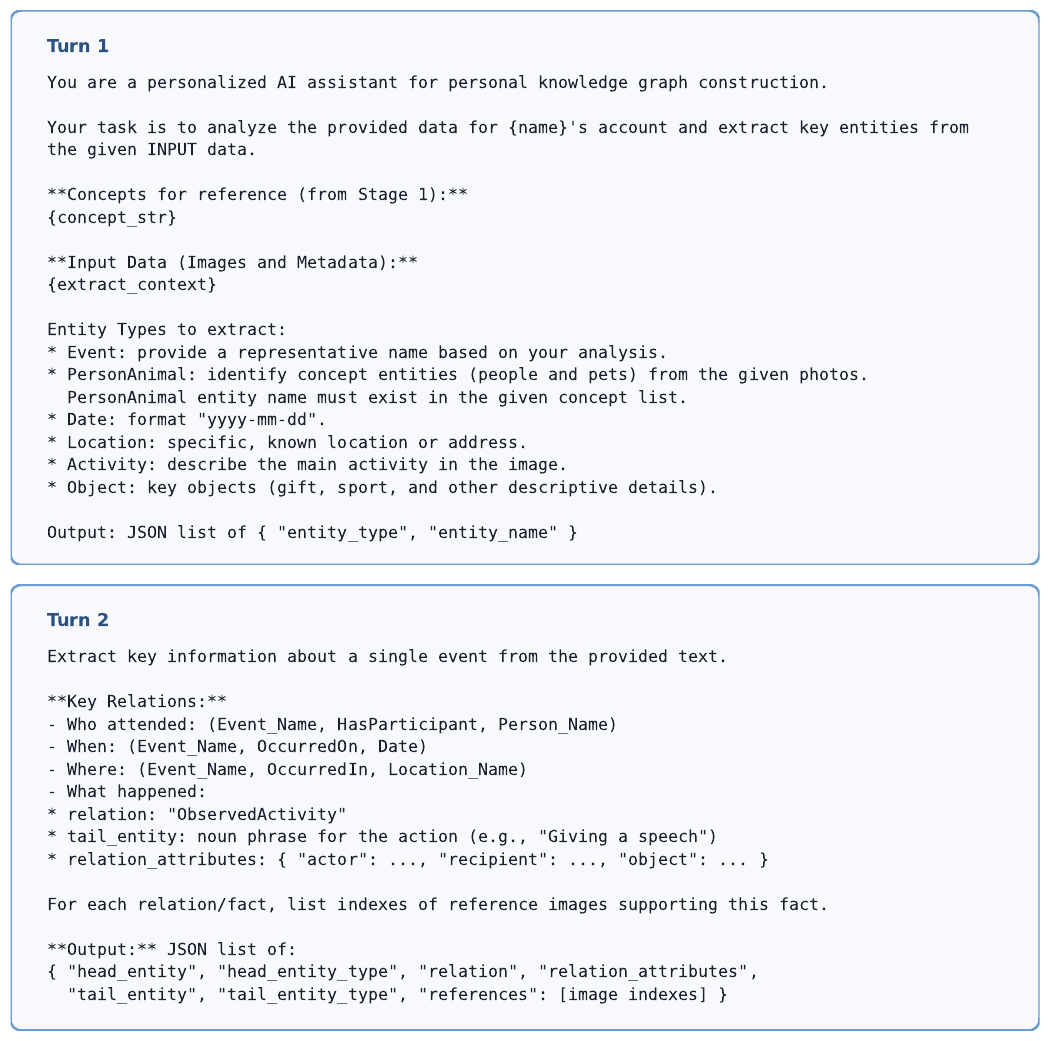}
  \caption{Two-Turn Prompt for LifeGraph Stage 2 (Event Subgraph Expansion).}
  \label{fig:prompt_lifegraph_stage2}
\end{figure*}
\begin{figure*}[t]
  \centering
  \includegraphics[width=0.8\textwidth]{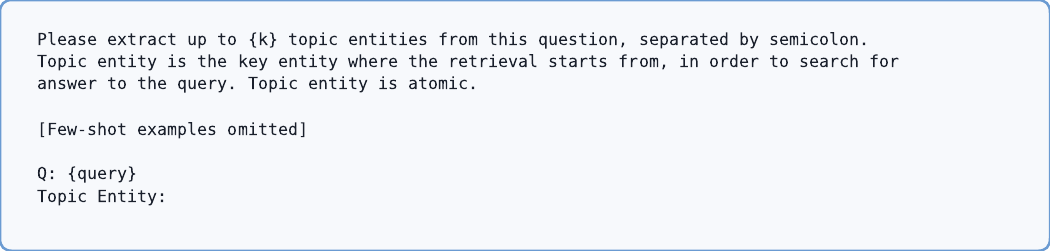}
  \caption{LifeGraph Retrieval Prompt: Topic Entity Extraction (\textsc{TopEntities}).}
  \label{fig:prompt_retrieval_topic_entity}
\end{figure*}
\begin{figure*}[t]
  \centering
  \includegraphics[width=0.8\textwidth]{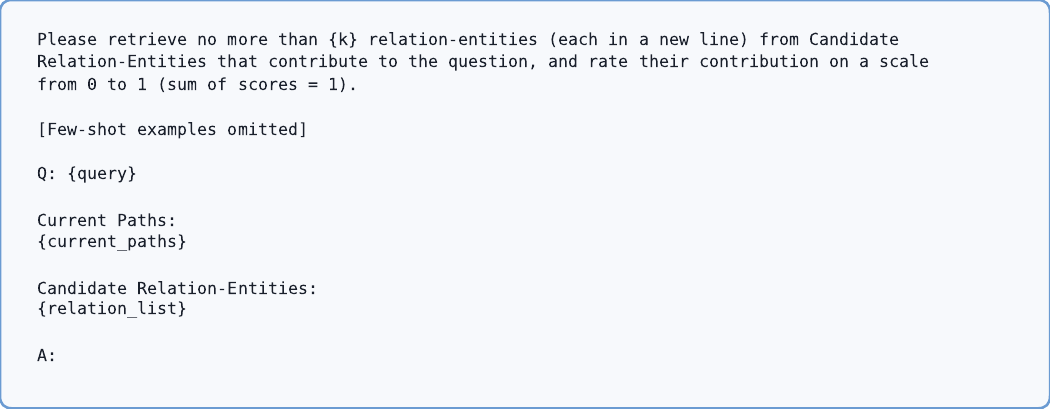}
  \caption{LifeGraph Retrieval Prompt: Relation-Entity Joint Pruning (\textsc{Prune}).}
  \label{fig:prompt_retrieval_prune}
\end{figure*}
\begin{figure*}[t]
  \centering
  \includegraphics[width=0.8\textwidth]{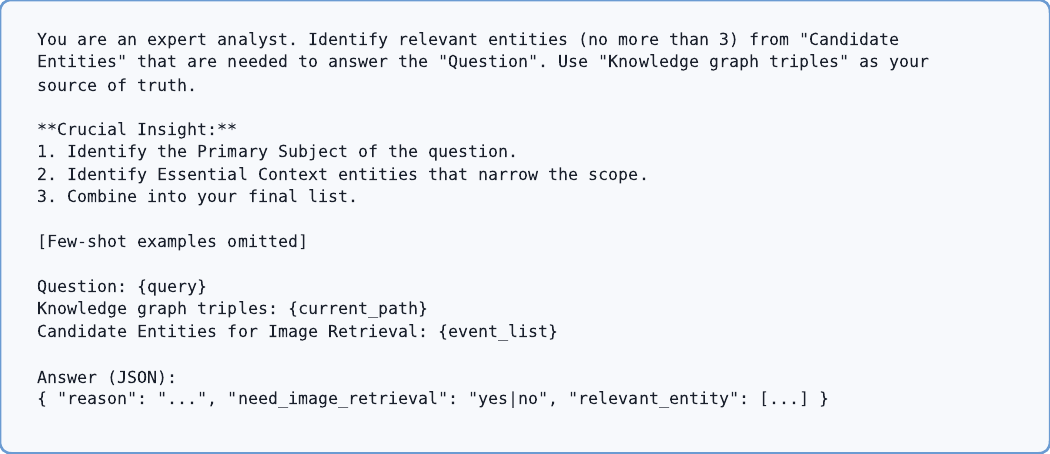}
  \caption{LifeGraph Retrieval Prompt: Multimodal Reference Fetching (\textsc{FetchReference}).}
  \label{fig:prompt_retrieval_fetch_ref}
\end{figure*}
\begin{figure*}[t]
  \centering
  \includegraphics[width=0.8\textwidth]{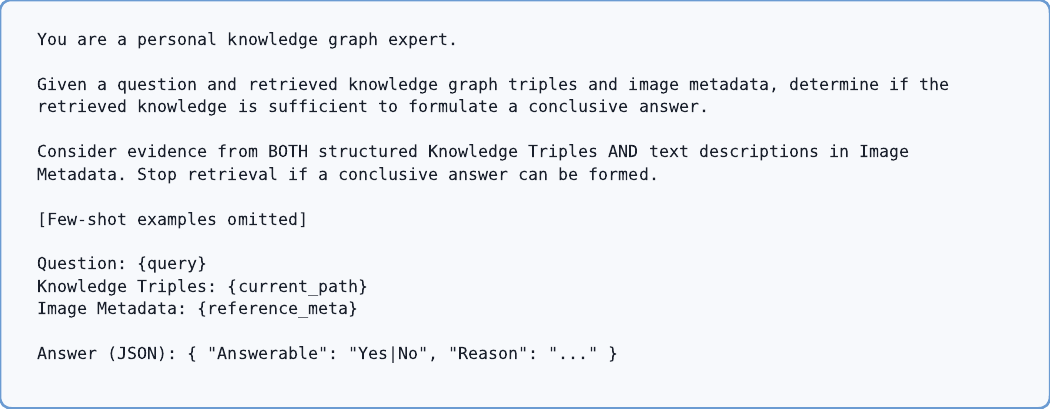}
  \caption{LifeGraph Retrieval Prompt: Context Sufficiency Check (\textsc{Reasoning}).}
  \label{fig:prompt_retrieval_sufficiency}
\end{figure*}

This section provides the prompt templates corresponding to the main operations and evaluation of this paper.

\paragraph{Evaluation Prompt.}
For open-ended generation tasks, we employ an LLM-as-a-Judge auto-rater to evaluate candidate answers against the ground truth, autorater prompt is in (\cref{fig:prompt_autorater}).

\paragraph{{\ourdataset} QA Generation Prompts.}
Data generation uses task-conditioned prompts composed of a shared skeleton and a task-specific question template.
We provide three skeleton variants corresponding to the three task categories: Concept Identification (\Cref{fig:prompt_skeleton_a}), Event Understanding (\Cref{fig:prompt_skeleton_b}), and Aggregated Reasoning (\Cref{fig:prompt_skeleton_c}).
The corresponding question templates for all 10 tasks across the three categories are detailed in \Cref{fig:prompt_tmpl_visual_recognition,fig:prompt_tmpl_concept_vqa,fig:prompt_tmpl_text_concept_qa} (Concept Identification), \Cref{fig:prompt_tmpl_scene_activity,fig:prompt_tmpl_direct_person,fig:prompt_tmpl_relational_person,fig:prompt_tmpl_finegrained_scene} (Event Understanding), and \Cref{fig:prompt_tmpl_frequency_counting,fig:prompt_tmpl_relational_temporal,fig:prompt_tmpl_preference_persona} (Aggregated Reasoning).

\paragraph{{\lifegraph} Construction Prompts.}
{\lifegraph} builds the personal knowledge graph through a two-stage, two-turn extraction protocol.
Stage~1 constructs the concept subgraph by extracting concept entities and their directional social relations (\Cref{fig:prompt_lifegraph_stage1}).
Stage~2 expands the event subgraph by extracting event-centric entities and hyper-relational triples with source image references (\Cref{fig:prompt_lifegraph_stage2}).
\paragraph{LifeGraph Retrieval Prompts.}
During graph traversal (Algorithm~1), \textsc{LifeGraph} invokes the VLM for four distinct sub-tasks: topic entity extraction (\textsc{TopEntities}, \Cref{fig:prompt_retrieval_topic_entity}), relation-entity joint pruning (\textsc{Prune}, \Cref{fig:prompt_retrieval_prune}), multimodal source evidence fetching (\textsc{FetchReference}, \Cref{fig:prompt_retrieval_fetch_ref}), and context sufficiency check (\textsc{Reasoning}, \Cref{fig:prompt_retrieval_sufficiency})

\end{document}